\documentclass[journal]{IEEEtran}
\usepackage{graphicx}
\usepackage{amsmath}
\usepackage{amssymb}
\usepackage{float}
\usepackage{caption}
\usepackage{subcaption} 
\usepackage{dblfloatfix}
\usepackage{wrapfig}
\usepackage{booktabs}
\usepackage{multirow}
\usepackage{balance}
\usepackage{xcolor}
\usepackage{algorithm}
\usepackage[noend]{algpseudocode}
\usepackage{lineno,hyperref}
\modulolinenumbers[5]
\usepackage{xcolor}
\usepackage{etoolbox}
\usepackage{url}
\usepackage{algorithmicx}
\usepackage [english]{babel}
\usepackage [autostyle, english = american]{csquotes}
\MakeOuterQuote{"}
\begin{document}
\title{Understanding Crowd Flow Movements Using Active-Langevin Model}
\author{Shreetam~Behera, \textit{Student Member, IEEE}, 
        Debi~Prosad~Dogra, \textit{Member, IEEE},~Malay Kumar Bandyopadhyay, and~Partha Pratim Roy, \textit{Member, IEEE}
\thanks{S. Behera and D. P. Dogra are with School 
of Electrical Sciences, Indian Institute of Technology Bhubaneswar,
Odisha, 752050 India e-mail: sb46@iitbbs.ac.in, dpdogra@iitbbs.ac.in.}
\thanks{M. K. Bandyopadhyay is with School of Basic Sciences,~Indian Institute of Technology Bhubaneswar, Odisha, 752050 India e-mail: malay@iitbbs.ac.in.}
\thanks{P. P. Roy is with Department of Computer Science and Engineering,~Indian Institute of Technology Roorkee, Uttarakhand, 247667~India e-mail: proy.fcs@iitr.ac.in.}
}

\IEEEtitleabstractindextext{%
\begin{abstract}
Crowd flow describes the elementary group behavior of crowds. Understanding the dynamics behind these movements can help to identify various abnormalities in crowds. However, developing a crowd model describing these flows is a challenging task. In this paper, a physics-based model is proposed to describe the movements in dense crowds. The crowd model is based on active Langevin equation where the motion points are assumed to be similar to active colloidal particles in fluids. The model is further augmented with computer-vision techniques to segment both linear and non-linear motion flows in a dense crowd. The evaluation of the active Langevin equation-based crowd segmentation has been done on publicly available crowd videos and on our own videos. The proposed method is able to segment the flow with lesser optical flow error and better accuracy in comparison to existing state-of-the-art methods.
\end{abstract}

}

\maketitle

\IEEEdisplaynontitleabstractindextext

\IEEEpeerreviewmaketitle

\section{Introduction}
\label{sec:intro}
In nature, collective behavior is one of the fundamental characteristics of different living organisms from bacteria to humans. Every collective movement of living organisms exhibit typical behavioral patterns indicating specific activities. For example, birds flock up together to drive out individuals of other species. Similarly, ants swarm together to drive larger pieces of food particles to their nests. In humans, collective motion can be seen in social events like rallies, parade, sports event, fairs and festivals. Understanding this kind of collective behavior can explain the cause of untoward incidents like stampede or other incidents that often cause loss of life and property. Researchers across various domains shown their interests in understanding the group behavior in humans~\cite{zhang2018physics}, thus making this as an inter-disciplinary field of research. 
In this paper, it has been shown how the amalgamation of physics-based method and computer vision techniques can be used to segment motion flows in densely crowded videos. The motion flows provide important cues about the crowd behavior that can be used for building systems to prevent crowd disasters.
\subsection{Related Work}
\label{ssec:lr}
Crowd motion flows can be very instrumental in describing group behavior in humans. Several literatures have explained crowd motion behavior on basis of physics and biological-inspired models \cite{kok2016crowd,zhang2018physics}. However, majority of such works primarily focus on physics-inspired models for crowd behavior analysis. These methods claim that the physics-based models can capture most of the dynamics of crowd motions. In physics-inspired models, the dominant flow in the crowd can be considered analogous to fluid. In an another approach, sparse human crowd can be considered similar to motion of gases~\cite{zhang2018physics}. Therefore, it is believed that the theory of fluid dynamics, statistical thermodynamics, concept of Brownian motion and many other physical concepts can only be employed after certain relaxations in the root models. In \cite{hughes2003flow}, Hughes pointed out the similarities between physics and the actual crowd. He described the crowd as a component of fluids from the physic's perspective. However, the concept is complex as interactions between individuals is far complicated than fluid particle interactions. Similarly, Vicsek et al. \cite{vicsek2012collective} have developed force-based models to describe collective motion in crowd. The collective motion is described in terms of velocity, orientation, correlation function and fluid dynamics. Helbing et al. \cite{helbing2007dynamics} have developed a physical model that uses density and pressure quantities to mark turbulent flows and stop-and-go phenomena in crowd. In \cite{alexiadis2004traffic}, the authors categorized crowd models as microscopic, mesoscopic and macroscopic. The macroscopic model considers crowd as a single unit comprising of all individuals. In the microscopic model, a person is considered as a fundamental unit of the crowd. The mesoscopic model is a combination of both macroscopic and microscopic model. However, the difference between the former two models is not crisp. In \cite{johansson2008crowd}, Johansson et al. have discussed about different dynamics of crowd that lead to various crowd safety issues.
The authors in \cite{fisher2009perfect} have used simple set of rules of interactions between neighboring particles in order to explain collective behavior with a special focus on group intelligence in human crowd. In \cite{lin2007crowd}, the authors have proposed a Particle Swarm Optimization model (PSO) to simulate crowd. The authors in \cite{bruno2011non} presented a physical modeling framework that describes the intelligent, non-local, and anisotropic behavior of pedestrians. In \cite{van2016lagrangian}, the Lagrangian (moving) coordinate system has been used for simulation and modeling of crowd flows. The authors in \cite{hoogendoorn2014continuum} have developed a multi-class continuum modeling based on social force model to simulate bilinear crowd flows. The authors in \cite{kulkarni2019sparse} have developed a dynamic variant of Vicsek model to study collective motion in panicked human crowds. 

\par Existing literatures \cite{kok2016crowd,zhang2018physics,junior2010crowd}                                                                                                                                                                                                                                                                                                                                                                          suggest that computer vision-based crowd behavior has been taken as an active area of research.~Automating the process using computer vision approaches results in better information fusion, thus leading to better accuracy and most importantly less error because of limited human interference \cite{zhang2018physics,junior2010crowd}. It has been reported that physics-based models can describe crowd well and coalescence of these models along with computer vision techniques can solve several crowd-related problems. 
\par Ali et al. in \cite{ali2007lagrangian} have considered human crowd as a fluid and developed a Lagrangian-based fluid dynamics framework for segmenting crowd flows in videos. The framework is also able to find out the instabilities of the crowd flows. However, the model developed is too complex and cannot handle dynamic background issues. A social force model-based method has been proposed in \cite{ji2017anomaly} to detect crowd anomaly at pixel and block levels. The authors in \cite{khan2016analyzing} have developed an algorithm to detect the sink modes and similar flow regions with similar physical motion patterns in the crowd. Mehran et al.~\cite{mehran2010streakline} have analyzed the motion flows by combining social force graph technique and streaklines in the crowd. A scene-structure-based force model is proposed in~\cite{ali2008floor} to detect individuals in high-density crowd by analyzing its static, dynamic, and boundary floor fields in videos. In \cite{wang2014high}, crowd is represented by fluid flow using a Lagrangian system and it uses streaklines in combination with potential functions used for segmentation as well as for abnormal behavior detection. The authors in \cite{wu2017bilinear} have analyzed the crowd behavior on the basis of bilinear interaction of curl and divergence of the motion flows. A spatio-temporal driving force-based group segementation scheme has been proposed in \cite{li2010group}. However, the model lacks view variant and its parameters need to be adjusted when the view changes. An adaptive human motion analysis and prediction method for understanding the motion patterns has been proposed in~\cite{chen2011adaptive}. The method explained in~\cite{solmaz2012identifying} identifies multiple crowd behaviors by performing stability analysis for dynamical systems, thus avoiding object detection, tracking, and training. However, their method fails to capture the randomness in the crowd. The authors in \cite{su2013large} have represented the crowd flow as a spatio-temporal viscous fluid field and proposed a method based on appearance and driven factor perspective to recognize the crowd behavior at a large scale. A density independent hydrodynamics model (DIHM) has been proposed in \cite{ullah2017density} to detect coherence regions in crowded scenes with ability to handle varying crowd density over time. However, the method does not segments well at finer level. In \cite{zhang2015flow}, texture-based method can only be used to represent crowd motion flows and background with varying textures. These textures of flow regions are used for people counting. The method proposed in \cite{lin2016diffusion} detects coherent regions in a crowded scene using a thermal diffusion model and time-series clustering. However, the method is not robust as coherent regions are lost when the motion and non-motion regions are merged over time. The authors in~\cite{wu2009crowd} have proposed a region growing segmentation scheme based on the translational domain for segmenting crowd flows. However, the method fails if the translational flow related to crowd regions is not local. The authors in \cite{kountouriotis2014agent} have developed a real-time agent-based model to understand crowd behavior on the basis of group dynamics and agent-based personality traits. However, the performance degrades when the number of agents increases. Zhou et al.~\cite{zhou2015learning} have represented the crowd collective behavioral patterns as a mixed model of dynamic pedestrian-agents. Since the model is a microscopic model, it fails to handle varying crowd density. In \cite{zhou2014CC}, the authors segment the collective regions in a crowd and measure the order of collectiveness of such regions. Fradi et al.~\cite{fradi2017crowd} have developed local descriptors to obtain semantic information and interactive sparse crowd behaviors in crowded scenarios. However, it is not clear how the method handles dense crowd. The technique proposed in \cite{basak2017developing} is an agent-based model that monitors and predicts evacuation ways in emergency situations. Though the results are good but the technique is computationally intensive. In \cite{wu2017crowd}, the motion trajectories have been analyzed using curl and divergence properties to identify different crowd movements. 
Lim et al.~\cite{lim2014detection} have developed a method to detect salient regions and instabilities in crowd by considering crowd as a dynamic system. In  \cite{behera2019estimation}, the authors have proposed a Langevin-based force model to segment crowd flows in dense crowded scenarios. However, the proposed method segments only linear crowd flow regions and the force model is based out of passive system of particles. The authors in \cite{cao2015large} and \cite{zhou2016spatial} have used Convolutional Neural Networks (CNN) to perform large scale crowd analysis. However, a large volume of labeled data is required for training whose preparation is a cumbersome task. A sparse representation-based scheme is proposed in \cite{yuan2016congested} for detecting anomalies in crowded scenes. Chaker et al. in~\cite{chaker2017social} have modeled the crowd using social force model and used an unsupervised approach for crowd anomaly detection. The authors in \cite{kruthiventi2015crowd} have considered the crowd motion flow as a Conditional Random Field to segment crowd motion flows in the videos. However, the method is not robust enough to handle to intersecting flows. In~\cite{chan2008modeling}, a dynamic mixture model of textures and expected-maximization (EM) algorithm are used to segment motion in traffic and crowd videos.

\subsection{Motivation and Contributions}
\label{ssec:contri}

It has been discussed in the previous section that majority of the existing crowd analysis frameworks fail to describe the random movements in crowd. The authors in \cite{kulkarni2019sparse} have described the crowd model in terms of modified Vicsek model. However, the model force components are not clearly explained with respect to crowd dynamics. 
Even though the authors in \cite{behera2019estimation} have described that the randomness of the crowd can be considered similar to Brownian motion of particles in fluid using a Langevin model, however, the model captures only linear flows. The model has been developed on the basis of passive system of particles where the particle's drift motion and confinement take place due to the random forces. The model does not explain the nature of the flows when the particles are part of an active system, i.e., when they have self-propelling energy to propagate in the fluid. Moreover, there are no literature available that describes the crowd motion using active Langevin model despite the same has been used in protein dynamics~\cite{schweitzer2007brownian} and slurry dynamics \cite{Bonkinpillewar2015}. This becomes a motivation for the present work to develop an active Langevin force model for crowd analysis. In~\cite{Bonkinpillewar2015} and \cite{mahapatra2017transitions}, the Viscek model have been used for slurry dynamics to understand the transitions in confined dense active-particle systems. This has motivated us to develop the present method by combining active Langevin force and other components of the force to obtain a model that can segment linear and non-linear motions in crowd. In this line, following research contributions have been made:
\begin{itemize}
\item[1] Formulation of a force model based on the active Langevin equation to understand flows in dense crowd scenarios by assuming motion points to be  analogous to self-driving particles in colloidal solutions.
\item[2] The above force model is then used with vision-based algorithm to segment linear and non-linear flows in dense crowd videos.
\end{itemize}

\par Rest of the paper is organized as follows. The underlying principle of Langevin equation is explained in Section~\ref{sec:basic_LE}. In Section~\ref{sec:proposed}, the proposed method is discussed. The results are presented in Section~\ref{sec:results}. Section~\ref{sec:Conclusions} concludes the paper and discusses a few possible future scopes in this research area.
 \section{Preliminary of Langevin Equation}
 \label{sec:basic_LE}
The random motion of a small particle (micron size) immersed in a fluid is known as Brownian motion. Early studies on this phenomenon are based on pollen grains, dust particles, and various other colloidal sized objects \cite{coffey2004langevin,langevin1908theorie}. Later on the theory of Brownian motion have been applied sucessfully to other phenomena~\cite{schweitzer2007brownian}. The fundamental equation based on Newtonian motion, which describes Brownian motion successfully, is known as the Langevin equation. This equation comprises of frictional forces and random forces. These forces are related to each others by the fluctuation-dissipation theorem. Considering the motion of spherical particle with mass $m$ and velocity $v$ in a fluid medium with viscosity $\eta$, (\ref{Eqn:le}) describes the Newton's equation of motion for the particle.
 \begin{equation}
 m\frac{\text{d}v(t)}{\text{d}t}= F_{total}(t)
  \label{Eqn:le}
\end{equation}
where $F_{total}(t)$ represents the overall instantaneous force experienced by the particle at time instant $t$.

The origin of this force is due to the interaction of the Brownian particle with the surrounding particles present in the medium. It is really hard to get an exact expression of $F_{total}(t)$. However, frictional force $-\gamma v$ primarily dominates $F_{total}(t)$, which is proportional to the velocity of the Brownian particle. According to Stokes law, friction coefficient $\gamma$ can be computed as presented in (\ref{Eqn:stokes}).
 \begin{equation}
\gamma= 6 \pi \eta a
  \label{Eqn:stokes}
\end{equation}
By substituting $F_{total}(t)$ in (\ref{Eqn:le}) with the frictional force, the Newton's equation of motion can be expressed now as represented in (\ref{Eqn:le_friction}),
 \begin{equation}
 m\frac{\text{d}v(t)}{\text{d}t}= -\gamma v
  \label{Eqn:le_friction}
\end{equation}
whose solution can be expressed in (\ref{Eqn:expvelocity}). Accordingly the velocity of the Brownian particle should decay to zero at longer time intervals. However, at thermal equilibrium at room temperature ($T$), the mean-squared velocity of the Brownian particle is $\langle v^{2}\rangle_{eq}=\frac{3 K_{B}T}{m}$. This indicates that the $F_{total}$ needs to be modified. The randomness of the trajectory of an individual particle indicates the existence of an additional random or fluctuating force $\xi(t)$.  Thus, the equation of motion is modified and described as in (\ref{Eqn:le_two}). 
 \begin{equation}
 v(t)= v(0)e^{-\frac{\gamma t}{m}}
  \label{Eqn:expvelocity}
\end{equation}
 \begin{equation}
 m\frac{\text{d}v(t)}{\text{d}t}= -\gamma v + \xi(t)
  \label{Eqn:le_two}
\end{equation}

The friction and noise both arise due to the interaction of the Brownian particle with its environment. The noise can be considered as a fluctuating force whose basic nature is given by its first and second moments as represented in (\ref{Eqn:moments}),
\begin{equation}
 \langle\xi(t)\rangle=0,~\langle\xi(t1)\xi(t2)\rangle=2B\delta(t1-t2)
   \label{Eqn:moments}
\end{equation}
 where $B$ is a measure of the strength of fluctuating force. There is no correlation between two distinct impacts occurring in two distinct intervals which is indicated by the delta function in time ($\delta(t)$). With the above properties, (\ref{Eqn:le_two}) can be solved for mean-squared velocity as presented in (\ref{Eqn:msq_velocity}),
  \begin{equation}
 \langle v^{2}(t)\rangle=v^{2}(0)e^{-\frac{2 \gamma t}{m}}+\frac{B}{\gamma m} (1-e^{-\frac{2 \gamma t}{m}}).
   \label{Eqn:msq_velocity}
\end{equation}
Over longer time intervals, the exponential terms in (\ref{Eqn:msq_velocity}) drop out and it confines to $\frac{B}{\gamma m}$. This ensures its equilibrium value to be $\frac{K_{B}T} {m}$ such that
\begin{equation}
B=\gamma K_{B} T.
\end{equation}
The above equation is known as fluctuating-dissipation theorem that establishes the relationship between the strength of the random force ($B$) with the magnitude of the frictional force ($\gamma v$). It represents the trade-off between the friction ($\gamma$) that tries to push the system to a completely "dead" state and the fluctuating force or noise force strives to keep the system "alive". This condition is necessary to maintain thermal equilibrium state at longer time intervals.
So far, the above discussion is limited for a free non-interacting Brownian particle. For confined (like Harmonic potential $V(x)=\frac{1}{2}mv_{0}^{2}x^{2}$) and interacting Brownian particles (like Lennard-Jones potential $V_{LJ}(x)=4\epsilon[(\frac{\sigma}{x})^{12}-(\frac{\sigma}{x})^{6}]$),~(\ref{Eqn:le_two}) can be modified further as expressed in (\ref{Eqn:passiveLE}),
\begin{equation}
m\frac{\text{d}v(t)}{\text{d}t}= -\gamma v(t) + F_{cons} + \xi(t)
  \label{Eqn:passiveLE}
\end{equation}
where $F_{cons}$ is the conservative force that can be expressed in terms of the potentials mentioned above ($F_{cons}=-\nabla V$ or $ F_{cons}=-\nabla V_{LJ}(x)$).

  The equation mentioned in (\ref{Eqn:passiveLE}) describes about passive Brownian system that should be in equilibrium over longer time intervals as the component forces try to balance out each others producing a unique stationary state given by Maxwell-Boltzmann distribution. All these types of undirected motions are considered as passive motion. This is because Brownian particle does not participate actively in this motion. On the other hand, active \footnote{The term "active" implies that the individuals particles or units move acquiring energy from the environment.} motion of Brownian particles depend on the supply of energy. In biological systems, this kind of active self-driven Brownian motion can be observed at different scales, ranging from cells \cite{alexiadis2004traffic} or simple micro-organisms to higher scale organisms like bird or fish. Finally human crowd movement can be demonstrated as active Brownian motion. Such type of motions can be applied to confined systems of particles (like human traffic flow in two dimension) that performs collective motion under far from equilibrium conditions. Now, one major question arises. How the known picture of passive motion mentioned in (\ref{Eqn:passiveLE}) needs to be modified to incorporate self or internal "activity" of the particles? Here, the main assumption is, there is additional inflow of energy that causes active motion and can be represented practically by negative dissipation in the direction of motion. Thus, it can be modeled by negative friction $\beta(r,v)$, i.e., represented as a function of position and velocity. Usually this kind of systems are far from equilibrium. The negative friction force does not obey fluctuation-dissipation relationship implying the system is considered as homogeneous in space implying $\beta(r,v) = \beta(v)$. This is considered as a frictional force applied to the component of the motion in the direction of the particle connecting vectors that helps particle moving together in the same direction. Thus, (\ref{Eqn:passiveLE}) is modified below and is represented as given in (\ref{Eqn:activeLE2}),
\begin{equation}
 m\frac{\text{d}v(t)}{\text{d}t}= -\gamma v(t)+(F_{conf}+F_{int})+F_{drift}+ \xi(t)
  \label{Eqn:activeLE2}
\end{equation}
where $F_{drift}=\beta v $. $F_{cons}$ is the combination of confinement force $F_{conf}$ and particle interaction force $F_int$, respectively.

\section{Proposed Crowd Flow Segmentation}
\label{sec:proposed}
This section describes the proposed algorithm that aimed at segmenting flows in densely crowded scenarios. 

\subsection{Keypoint Extraction}
\label{subsec:ke}
The proposed algorithm works on a temporal window of size, say $W$ frames. Over the first two frames, absolute difference is computed. This step retains all motion regions in the scene. Next, Features from Accelerated Segment Test (FAST) detector \cite{rosten2010faster} is applied to detect important keypoints in the crowded scene. Applying FAST detector to difference image has two advantages. Firstly, it retains only the motile points. Secondly, computations are performed only on the keypoints reducing the computational time. These keypoints are then fed to the Lucas Kanade Optical flow process \cite{lucas1981iterative} for tracking in the subsequent frames within the window $W$. The magnitudes of the keypoints are calculated using (\ref{Eqn:magnitude}) and (\ref{Eqn:orientation}), respectively. The detailed implementation is explained in the Algorithm~\ref{algo:keypoints}.
\begin{equation}
v = \sqrt{|v_{x}|^2+|v_{y}|^2}
\label{Eqn:magnitude}
\end{equation}
\begin{equation}
\theta=\arctan {(|v_y|/|v_x|)}
\label{Eqn:orientation}
\end{equation}
\begin{algorithm}
\scriptsize
\textbf{Input:} $W (f_{1}, f_{2})$ = First two frames of the a Temporal Window $W$. \\
\textbf{Output:} $K$ = Set of keypoints with $v_{x}, v_{y}, m, \theta, Q$ as features of each keypoint in $K$.
\begin{algorithmic}[1]
\caption{Keypoint extraction}
\label{algo:keypoints}
\State Compute absolute difference image $diff$=$|f_{2}-f_{1}|$
\State Compute FAST keypoints ($K_{f}$) on the $diff$ image.
\State Calculate ${v_{x}, v_{y}}$ using Lucas Kanade Optical Flow method$(f_{1}, f_{2})$.
\State Calculate $ M and~\theta$ using (\ref{Eqn:magnitude}) and (\ref{Eqn:orientation}).
\State Compute Q by quantizing $\theta$ into $b$ bins in the range of $0$-$2\pi$.
\end{algorithmic}
\end{algorithm}
\subsection{Active Langevin Force Model}
\label{subsec:le}
Initially, the formulation of active Langevin model is discussed. The detected keypoints as discussed in the previous section, are considered to be in motion that constitute the overall flow in a crowd. The motion keypoints can be considered as self-propelling particles as in active systems moving with certain drift energy. Similar to \cite{behera2019estimation}, the inertial force ($F_{inertial}$) constitutes of three different forces as given in (\ref{Eqn:ForceInertia}).
\begin{equation}
{F_{inertial}=~F_{external}+F_{active}+F_{disturbance}}
\label{Eqn:ForceInertia}
\end{equation}
The first term in the right hand side of (\ref{Eqn:ForceInertia}) represents the viscous force similar to friction force in (\ref{Eqn:passiveLE}). The second force ($F_{active}$) represents  the combination of interaction potential resulted due to the interaction among the particles and the drift force responsible for self-driving of the particle as  mentioned in (\ref{Eqn:ActiveForce}). The third term is a random force resulted because of random noise and disturbances. Now, the reformulated Langevin force equation for the $i^{th}$ particle in $n$ dimensions can be presented as in~(\ref{Eqn:activeLE}),
\begin{equation}
 m_{i}\frac{\text{d}v_{i,n}(t)}{\text{d}t}= -\gamma v_{i,n}(t)+F_{active_{i,n}}+  \xi(t)
  \label{Eqn:activeLE}
\end{equation}
where $m$ is the mass of the particle, $v_{i,n}(t)$ represents the velocity of the $i^{th}$ particle in $n_{th}$ direction, $-\gamma$ represents the viscosity coefficient, and the $\xi(t)$ represents a random force. The term $F_{active_{i,n}}$ further consists of interaction potential and drift force as represented in (\ref{Eqn:ActiveForce}),
\begin{equation}
 F_{active_{i,n}}= -\nabla{U_{i,n}}+F_{drift_{i,n}}
  \label{Eqn:ActiveForce}
\end{equation}
where $-\nabla{U_{i,n}}$ represents interaction potential, $U_{i,n}$ is the force potential, and $F_{drift_{i,n}}$ represents the drift force experienced by the particle.
\subsection{Flow Segmentation Method}
\label{ssec:implementation}
The crowd movements are considered as $2D$ translational movements. Thus, the value of $n$ is assumed to be $2$. Equation (\ref{Eqn:activeLE}) is further solved w.r.t. change in time ($\Delta{t}$) to compute the corresponding velocities and positions of the particle in the next time frame. 
\begin{equation}
v_{new_{i,n}}=v_{old_{i,n}}-\gamma v_{new_{i,n}}{\Delta}t+F_{active}{\Delta}t+\xi(t){\Delta}t
\label{Eqn:VelPred}
\end{equation}
The above equation in (\ref{Eqn:VelPred}) represents the predicted velocity of the particle in the next time frame. Similarly, the position of the particle is computed as mentioned in (\ref{Eqn:PosPred}),
\begin{equation}
r_{new_{i,n}} = r_{old_{i,n}}+v_{new_{i,n}}{\Delta}t
\label{Eqn:PosPred}
\end{equation}

where $\Delta$t is the increment in time. In the above equations, the mass of each particle is set to unity for consistency and since the operations are performed in consecutive frames, $\Delta t$ is taken as unity.
The forces mentioned in (\ref{Eqn:ForceInertia}) can be computed as follows:
\begin{itemize}
\item \textbf{Estimation of Viscous Force:} The viscous force can be calculated as the product of the particle velocity and viscosity of the particle and its neighbors. Viscosity is calculated as mentioned below in (\ref{Eqn:viscosity}),
\begin{equation}
\gamma = 1-\frac{1}{k_{n}*max(|r_{i}-r_{k}|)} \sum_{k=1}^{k_{n}}|r_{i}-r_{k}|
\label{Eqn:viscosity}
\end{equation}
where $|.|$ represents the distance between the two particles, $r_{i}$ represents the position of the $i^{th}$ particle, $r_{k}$ position of the $k^{th}$ neighbor of the $i^{th}$ particle, and $k_{n}$ represents the total number of neighbors surrounding the considered particle. Thus, the viscous force is calculated as (\ref{Eqn:viscousForce}).
\begin{equation}
F_{external}=-\gamma v_{i}(t)
\label{Eqn:viscousForce}
\end{equation}
\\
\item \textbf{Estimation of Active Force:} As mentioned in (\ref{Eqn:ActiveForce}), this force has two parts namely the particle interaction potential and the drift force. The interaction potential can be considered as the average interactions of the $i^{th}$ particle with its neighbors as presented in (\ref{Eqn:FinterP}),
\begin{equation}
F_{interaction}=-\mu (v_{i}(t)-v_{rel}(t))
\label{Eqn:FinterP}
\end{equation}
where $\mu$ represents the interaction coefficient known as coordination coefficient that happens due to the interactions of the $i^{th}$ particle and its neighbors as presented in (\ref{Eqn:muCalc}), $v_{avg}(t)$ is the average particle velocity and $v_{rel}(t)$ is the relative velocity,
\begin{equation}
\mu=\frac{ v_{rel}(t) }{ v_{avg}(t) }
\label{Eqn:muCalc}
\end{equation}
where $v_{avg}(t)$ is represented as in (\ref{Eqn:vavg}), and $v_{rel}(t)$ is expressed as in (\ref{Eqn:relV}).
\begin{equation}
v_{avg}(t)=\frac{1}{k_{n}}\sum_{k=1}^{k_{n}}v_{k}(t)
\label{Eqn:vavg}
\end{equation}
\begin{equation}
v_{rel}(t)=\frac{\sum_{k=1}^{k_{n}}W_{ik}(||r_{i}-r_{j}||,h)v_{k}(t)}{\sum_{k=1}^{k_{n}}W_{ik}(||r_{i}-r_{j}||,h)}
\label{Eqn:relV}
\end{equation}
In the aforementioned formulation, $h$ represents the radius upto which the potential influence is experienced, $k_{n}$ represents the number of neighbors across the $i^{th}$ particle, and $W_{ik}(||r_{i}-r_{j}||, h)$ is represented as the Gaussian weight function as described in (\ref{Eqn:gaussW}).
\begin{equation}
W_{ik}= 
    \begin{cases}
    e^{-\frac{||r_{i}-r_{k}||^2}{h^2}}, & \text{if}\ \frac{||r_{i}-r_{j}||}{h} <=1 \\
      0, & \text{otherwise}
    \end{cases}    
\label{Eqn:gaussW}
\end{equation}
The drift force can now be calculated as given in (\ref{Eqn:driftForce}), 
\begin{equation}
F_{drift_{i}}=\beta v_{i}(t)
\label{Eqn:driftForce}
\end{equation}
where $\beta$ is the self-propelling coefficient. The sum of $F_{interaction}$ and $F_{drift}$ constitutes the active force $F_{active}$.
\\
\item \textbf{Estimation of Random Force:} This force is taken as the force generated randomly at any point of time due to disturbances.
\end{itemize}
\par
\begin{figure*}[hbt!]
	\centering	
    \includegraphics[scale=1.0,width=0.9\textwidth]{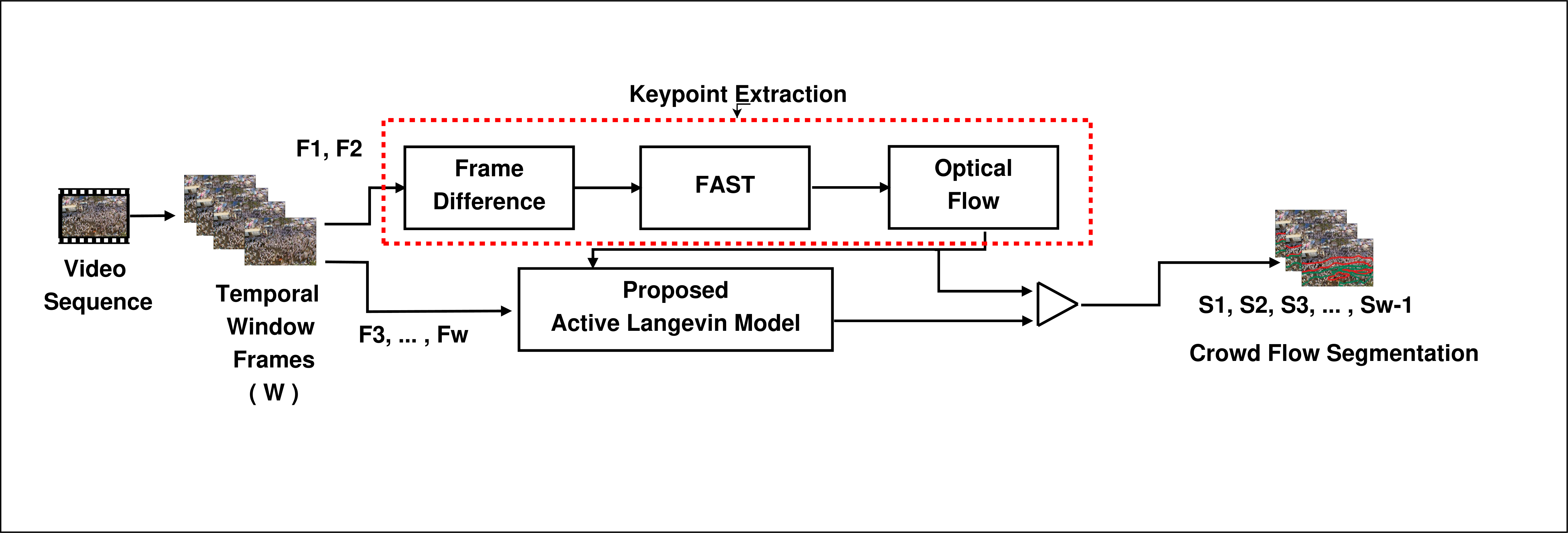}
    \caption{Block diagram representing proposed crowd flow segmentation using active Langevin model. Inside the red dotted box, the keypoint extraction scheme is shown. For a temporal window $W$, the first two frames are used for keypoint extraction. These keypoints are then used to segment the crowd motion flows in the remaining frames of the window using active Langevin model.}  
    \label{fig:BD}
\end{figure*}

 The active Langevin equation model is applied to each and every keypoints in order to obtain velocity and position across $x$ and $y$-axes, respectively for next frame in the window $W$. Again the keypoints obtained for the current frame are used to compute the keypoints in the next frame within the window and the process continues till the last frame of the window. The flow segmentation process is explained in Algorithm (\ref{algo:computeLE}) and illustrated in the Figure (\ref{fig:BD}), respectively.

\begin{algorithm}
\scriptsize
\textbf{Input:} $F (f_{1}, f_{2}, f_{3}, ...~, f_{T})$ = Video sequence with $T$ number of frames, $|W|$ = Size of Window, $beta$, $b$ = Quantization bins. \\
\textbf{Output:} $G_{s}$ = Motion flow segmented maps, where $s$= $|W|-1$
\begin{algorithmic}[1]
\caption{Crowd flow segmentation using active Langevin Model}
\label{algo:computeLE}
\State Initialize $m$~=~$\frac{T}{|W|}$.  
\For {i = 1 to m}
\State $W_{i}={f_{p+1}, f_{p+2}, ... ,f_{p+|W|}}$, where $p = i*|W|$
\State Extract keypoints $K$ using \textbf{algorithm (\ref{algo:keypoints})}.
	\For {j = 3 to $|W|$}
		\State Using \textbf{(\ref{Eqn:VelPred})} and \textbf{(\ref{Eqn:PosPred})}, estimate the new velocities and positions of the particles present in $K$.
	\EndFor
\EndFor
\end{algorithmic}
\end{algorithm}
\section{Results and Discussions}
\label{sec:results}
This section discusses about the datasets used for evaluation of the proposed scheme, followed by experiments related to parameter estimation of the force model.
\subsection{Datasets}
\label{ssec:datasets}
In this work, two datasets have been used for the evaluation of the proposed method. One of the datasets is publicly available~\cite{ali2007lagrangian} and second one is our own dataset containing video recordings of Rath Yatra (Cart Festival) that happens each year in India at Puri in the state of Odisha. From these datasets, a few videos with varying crowd densities have been selected for experimentation. The details of the videos from the two datasets are presented in Table \ref{table:datasets}.
\begin{table}[H]
\scriptsize
\centering
\caption{Videos from the two datasets used for evaluation of the proposed method}\label{table:datasets}
\resizebox{0.47\textwidth}{!}{%
\begin{tabular}{|l|c|c|} \hline
\#Dataset & Types of Motion & \begin{tabular}[c]{@{}c@{}}Significant Crowd \\ behavior\end{tabular} \\ \hline
Marathon-I~\cite{ali2007lagrangian} & \begin{tabular}[c]{@{}c@{}}Linear, unidirectional\\  crowd movements\end{tabular} & \begin{tabular}[c]{@{}c@{}}People running \\ in one direction\end{tabular} \\ \hline
Marathon-III~\cite{ali2007lagrangian} & \begin{tabular}[c]{@{}c@{}}Non-Linear, multidirectional\\  crowd movements\end{tabular} & \begin{tabular}[c]{@{}c@{}}People running\\ in elliptical path\end{tabular} \\ \hline
Fair~\cite{ali2007lagrangian} & \begin{tabular}[c]{@{}c@{}}Bilinear, mixing \\ crowd movements\end{tabular} & \begin{tabular}[c]{@{}c@{}}People moving in \\ two different directions\end{tabular} \\ \hline
Rath Yatra-I & \begin{tabular}[c]{@{}c@{}}Linear, mixing \\ crowd movements\end{tabular} & \begin{tabular}[c]{@{}c@{}}People pulling \\ the cart in one direction\end{tabular} \\ \hline
\end{tabular}%
}\end{table}
\subsection{Parameter Estimation}
\label{ssec:paraE}
\begin{figure}[hbt!]
	\centering	
	\includegraphics[width=9.50 cm,height=4.0 cm]{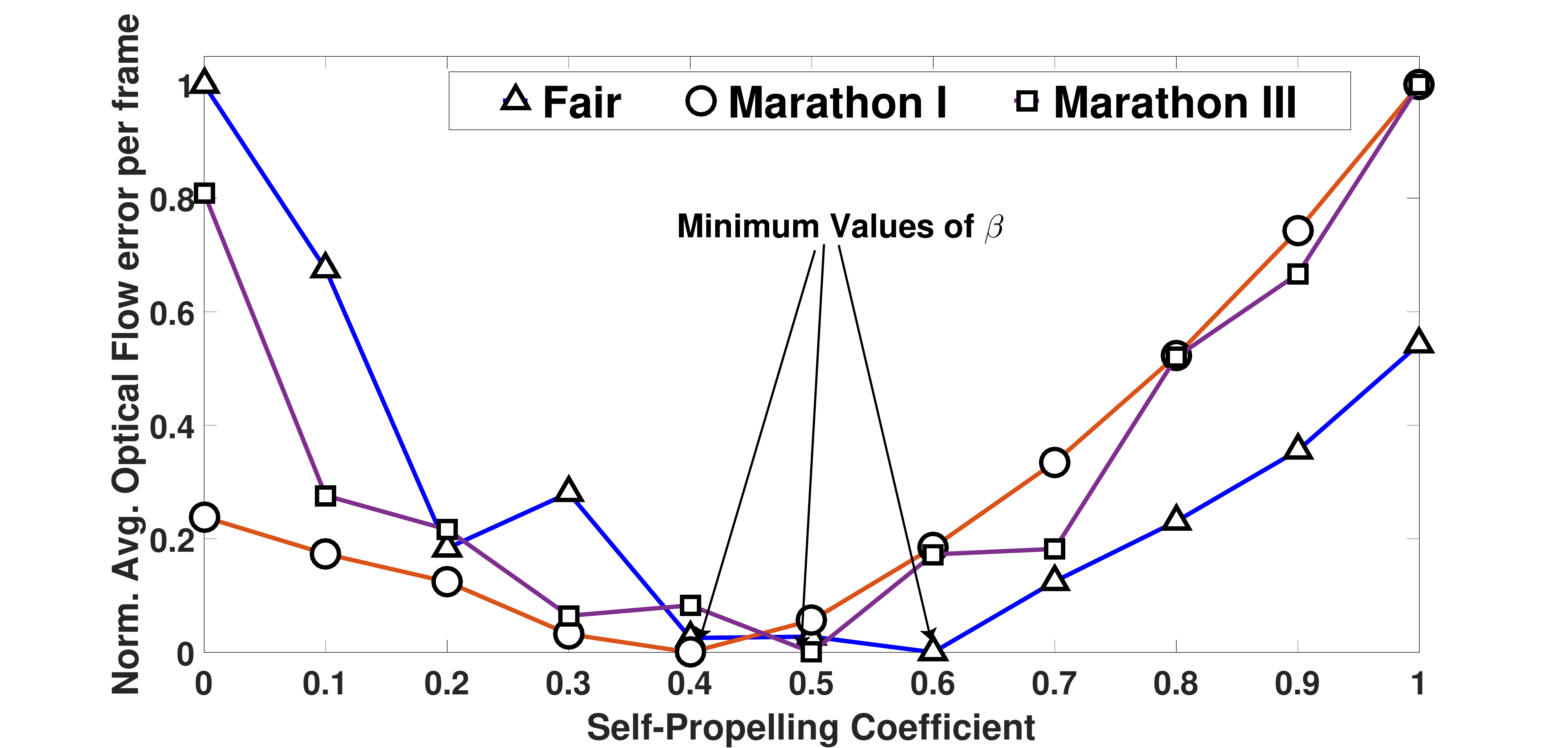}
    \caption{Graph showing how average optical flow error varies with respect to self-propelling coefficient ($\beta$).}  
    \label{fig:betaEstimation}
\end{figure}

The equations (\ref{Eqn:VelPred}) and (\ref{Eqn:PosPred}) described earlier have parameters such as viscosity coefficient ($\gamma$), 
coordination coefficient ($\mu$) and self-propelling coefficient($\beta$). The viscosity coefficient and coordination coefficient are calculated during the segmentation process itself. However, self-propelling coefficient needs to be given as input. Therefore, an experiment has been conducted to find optimal value of $\beta$ with respect to average optical flow error generated during the process. The experiment has been carried out on various videos with different movements. Videos with linear, non-linear, and crowd mixing movements have been considered. For each value of $\beta$, the normalized average optical flow error per frame is obtained. For each video, $\beta$ with minimum error is chosen. In order to obtain a uniform value of $\beta$ for all videos, the average value of all chosen minimum $\beta$ is considered as the final value of $\beta$. The graphs associated with this experiment are illustrated in Figure~\ref{fig:betaEstimation} and the minimum values of $\beta$ are presented in Table \ref{table:betaEstimation}. The average $\beta$ value has been found to be $0.5$ that has been kept fixed for all other videos.
\begin{table}[H]
\centering
\scriptsize
\caption{Minimum values of $\beta$ for different videos obtained from the graphs displayed in Figure \ref{fig:betaEstimation}}
\label{table:betaEstimation}

\resizebox{0.275\textwidth}{!}{%
\begin{tabular}{|l|c|}
\hline
\#Videos & $\beta_{minimum}$ \\ \hline
Fair & 0.6 \\ \hline
Marathon-I & 0.4 \\ \hline
Marathon-III & 0.5 \\ \hline
\textbf{Average} & \textbf{0.5} \\ \hline
\end{tabular}%
}
\end{table}
\subsection{Segmentation Results}
\label{ssec:segment}
In this section, the segmentation results obtained using the proposed flow segmentation method and comparisons with latest physics-based model for crowd segmentation, are discussed. The obtained segmentation maps are compared with the ground truths. The ground truths have been prepared by manually marking the significant flow regions in the frames of the video. The comparisons are done with the recent physics-based model for collective motion in crowds \cite{kulkarni2019sparse}, \cite{mahapatra2017transitions} and along with the hydrodynamics-based model proposed in \cite{ullah2017density}, respectively. Intersection over Union (IoU) also known as the Jaccard's coefficient has been used for evaluation of segmented maps with respect to ground maps as represented in (\ref{Eqn:accuracy}),

\begin{equation}
\label{Eqn:accuracy}
Accuracy=\frac{Area(S_{w} \cap G_{T})}{Area(S_{w} \cup G_{T})}
\end{equation}
where $S_{w}$ is the segmented image and $G_{T}$ is the ground truth image. 

\par
\begin{figure*}[hbt!]     
    \centering
   \captionsetup[subfigure]{labelformat=empty}
\begin{subfigure}[t]{0.156\textwidth}
\includegraphics[scale=0.1,width=0.8\textwidth]{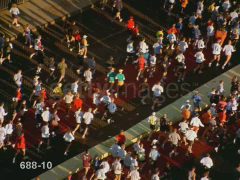}
        \caption{a} 
        \label{fig:Original68816}
    \end{subfigure}
       \begin{subfigure}[t]{0.156\textwidth}
\includegraphics[scale=1.0,width=0.8\textwidth]{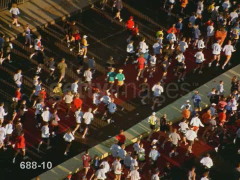}
        \caption{b} 
        \label{fig:Original68817}
    \end{subfigure}
       \begin{subfigure}[t]{0.156\textwidth}
        \includegraphics[scale=0.1,width=0.8\textwidth]{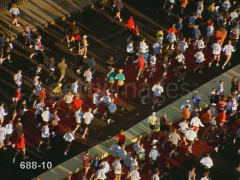}
        \caption{c} 
        \label{fig:Original68818}
        \end{subfigure}
          \begin{subfigure}[t]{0.156\textwidth}
        \includegraphics[scale=0.1,width=0.8\textwidth]{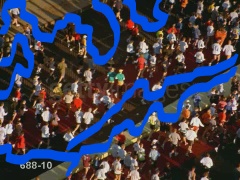}
        \caption{d} 
        \label{fig:GT68816}
    \end{subfigure}
       \begin{subfigure}[t]{0.156\textwidth}
        \includegraphics[scale=0.1,width=0.8\textwidth]{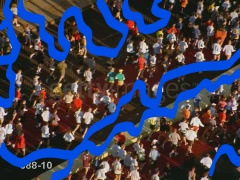}
        \caption{e} 
        \label{fig:GT68817}
    \end{subfigure}
       \begin{subfigure}[t]{0.156\textwidth}
        \includegraphics[scale=0.1,width=0.8\textwidth]{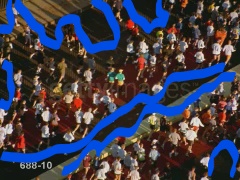}
        \caption{f} 
        \label{fig:GT68818}
        \end{subfigure}
        \begin{subfigure}[t]{0.156\textwidth}
        \includegraphics[scale=0.1,width=0.8\textwidth]{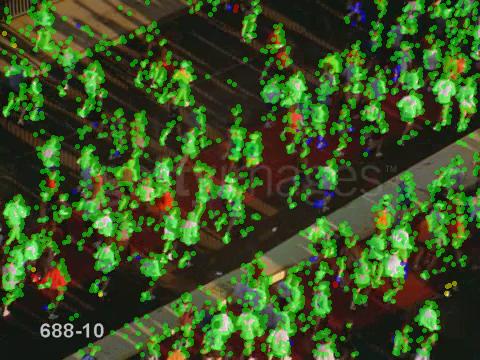}
        \caption{g} 
        \label{fig:SD68816}
    \end{subfigure}
       \begin{subfigure}[t]{0.156\textwidth}
        \includegraphics[scale=0.1,width=0.8\textwidth]{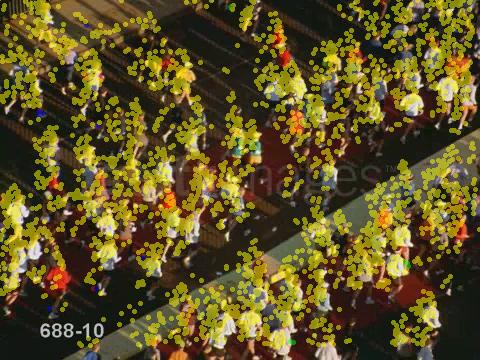}
        \caption{h} 
        \label{fig:SD68817}
    \end{subfigure}
       \begin{subfigure}[t]{0.156\textwidth}
        \includegraphics[scale=0.1,width=0.8\textwidth]{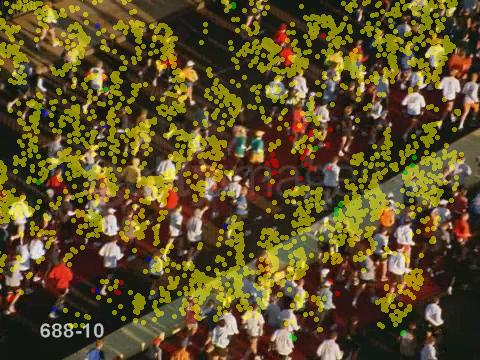}
        \caption{i} 
        \label{fig:SD68818}
        \end{subfigure}
         \begin{subfigure}[t]{0.156\textwidth}
        \includegraphics[scale=1.0,width=0.8\textwidth]{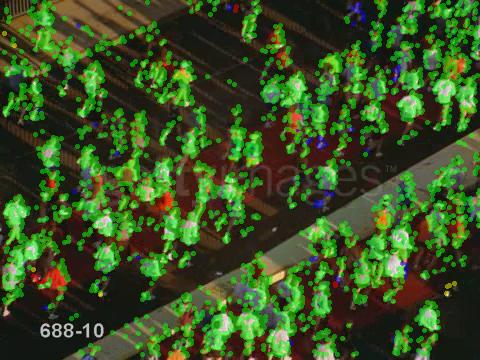}
        \caption{j} 
        \label{fig:AK68816}
    \end{subfigure}
       \begin{subfigure}[t]{0.156\textwidth}
        \includegraphics[scale=0.1,width=0.8\textwidth]{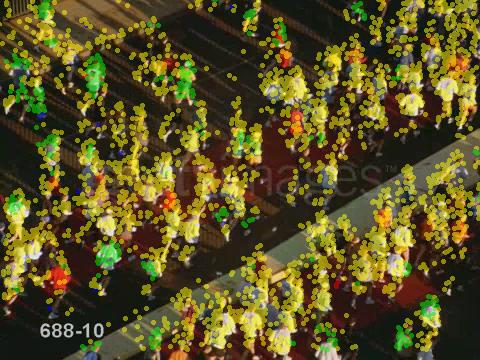}
        \caption{k} 
        \label{fig:AK68817}
    \end{subfigure}
       \begin{subfigure}[t]{0.156\textwidth}
        \includegraphics[scale=0.1,width=0.8\textwidth]{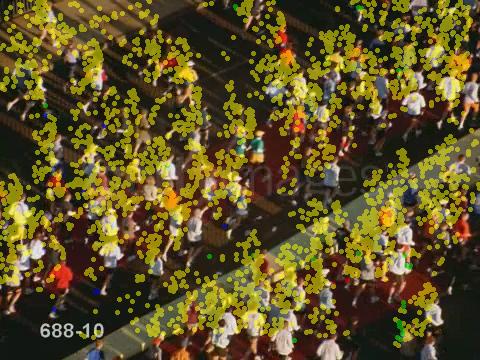}
        \caption{l} 
        \label{fig:AK68818}
        \end{subfigure}
        \begin{subfigure}[t]{0.156\textwidth}
        \includegraphics[scale=0.1,width=0.8\textwidth]{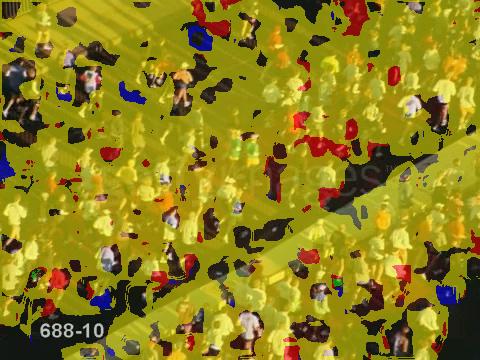}
        \caption{m} 
        \label{fig:DIHM68816}
        \end{subfigure}
               \begin{subfigure}[t]{0.156\textwidth}
        \includegraphics[scale=0.1,width=0.8\textwidth]{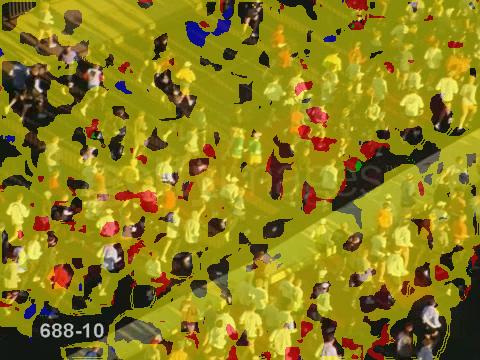}
        \caption{n} 
        \label{fig:DIHM68817}
        \end{subfigure}
               \begin{subfigure}[t]{0.156\textwidth}
        \includegraphics[scale=0.1,width=0.8\textwidth]{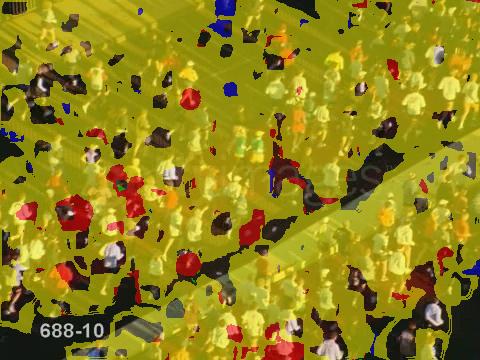}
        \caption{o} 
        \label{fig:DIHM68818}
        \end{subfigure}
        \begin{subfigure}[t]{0.156\textwidth}
        \includegraphics[scale=1.0,width=0.8\textwidth]{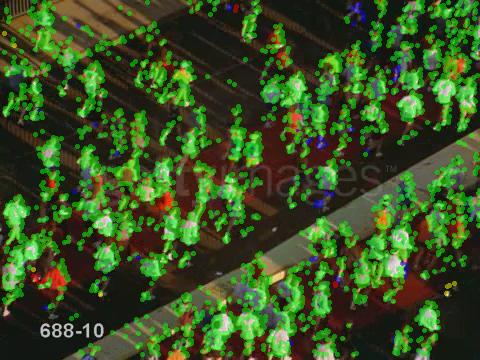}
        \caption{p} 
        \label{fig:ALE68816}
    \end{subfigure}
       \begin{subfigure}[t]{0.156\textwidth}
        \includegraphics[scale=0.1,width=0.8\textwidth]{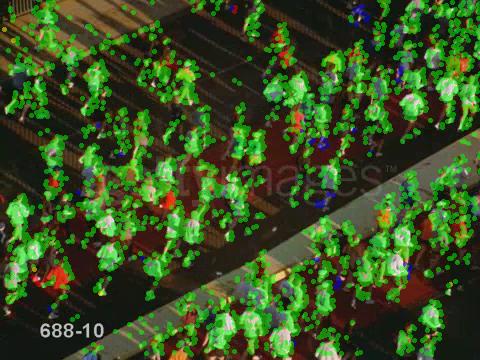}
        \caption{q} 
        \label{fig:ALE68817}
    \end{subfigure}
       \begin{subfigure}[t]{0.156\textwidth}
        \includegraphics[scale=0.1,width=0.8\textwidth]{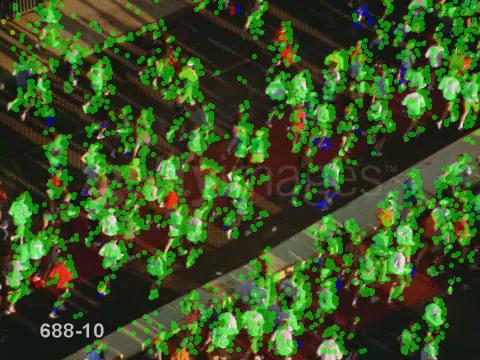}
        \caption{r} 
        \label{fig:ALE68818}
        \end{subfigure}
      \caption{($a$-$c$) Original Frames~(16-18) of the Marathon-I video, ($d$-$f$) Ground Truth Frames, ($g$-$i$) represent segmented outputs obtained using method proposed in~\cite{mahapatra2017transitions}, ($j$-$l$) represent segmentation outputs of  method~\cite{kulkarni2019sparse}, ($m$-$o$) represent segmentation outputs of method~\cite{ullah2017density}, and  ($p$-$r$) represent segmentation outputs of the proposed method, respectively. (Best viewed in color)}

\label{fig:688}
\end{figure*}

\begin{figure*}[hbt!]     
    \centering
    \captionsetup[subfigure]{labelformat=empty}
\begin{subfigure}[t]{0.154\textwidth}
        \includegraphics[scale=0.1,width=0.8\textwidth]{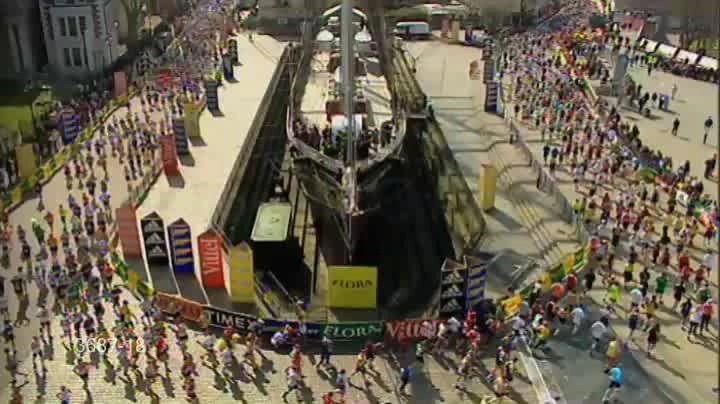}
        \caption{a} 
        \label{fig:3687_00359}
    \end{subfigure}
       \begin{subfigure}[t]{0.154\textwidth}
        \includegraphics[scale=0.1,width=0.8\textwidth]{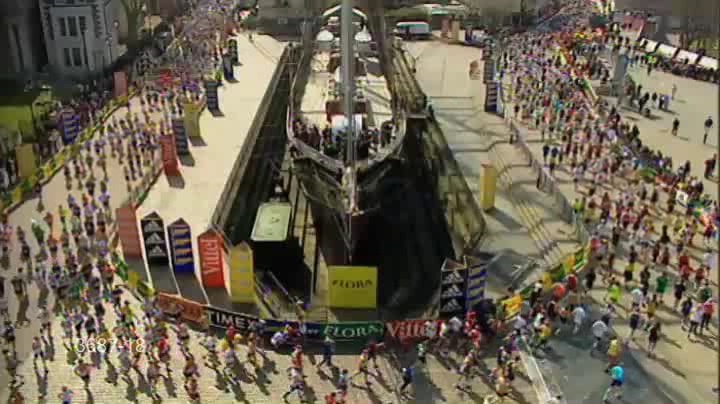}
        \caption{b} 
        \label{fig:3687_00361}
    \end{subfigure}
       \begin{subfigure}[t]{0.154\textwidth}
        \includegraphics[scale=0.1,width=0.8\textwidth]{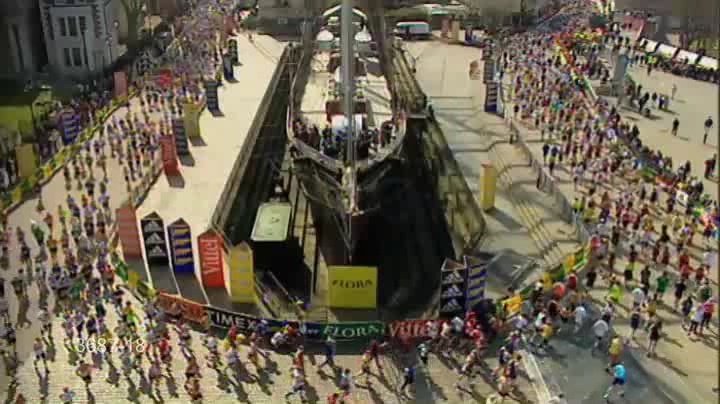}
        \caption{c} 
        \label{fig:3687_00362}
        \end{subfigure}
          \begin{subfigure}[t]{0.154\textwidth}
        \includegraphics[scale=0.1,width=0.8\textwidth]{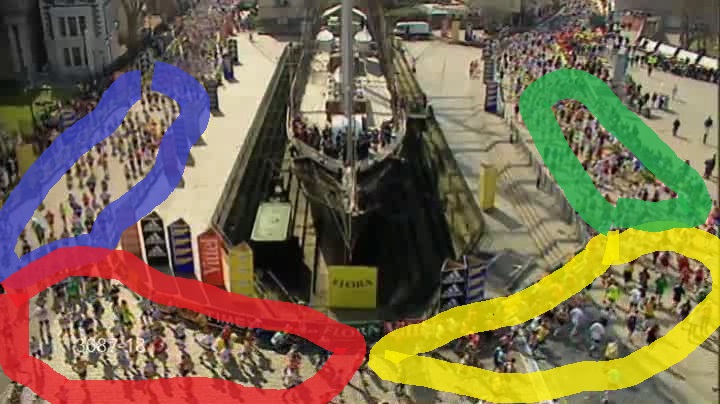}
       \caption{d} 
        \label{fig:GT368761}
    \end{subfigure}
       \begin{subfigure}[t]{0.154\textwidth}
        \includegraphics[scale=0.1,width=0.8\textwidth]{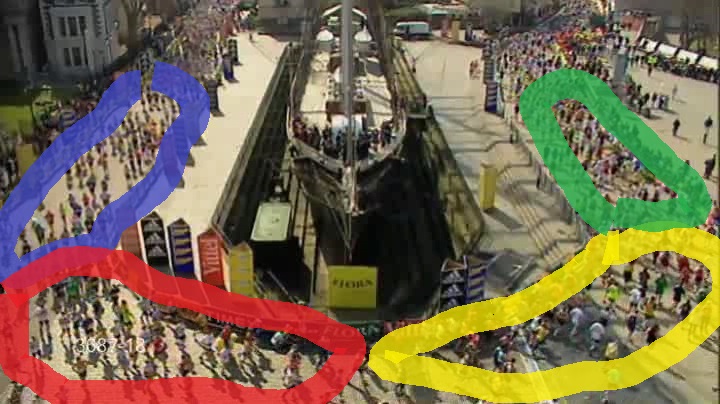}
        \caption{e} 
        \label{fig:GT368762}
    \end{subfigure}
       \begin{subfigure}[t]{0.154\textwidth}
        \includegraphics[scale=0.1,width=0.8\textwidth]{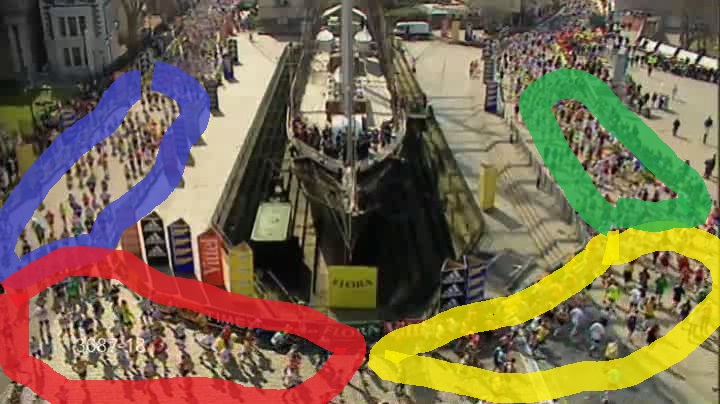}
        \caption{f} 
        \label{fig:GT368763}
        \end{subfigure}
                   \begin{subfigure}[t]{0.154\textwidth}
        \includegraphics[scale=0.1,width=0.8\textwidth]{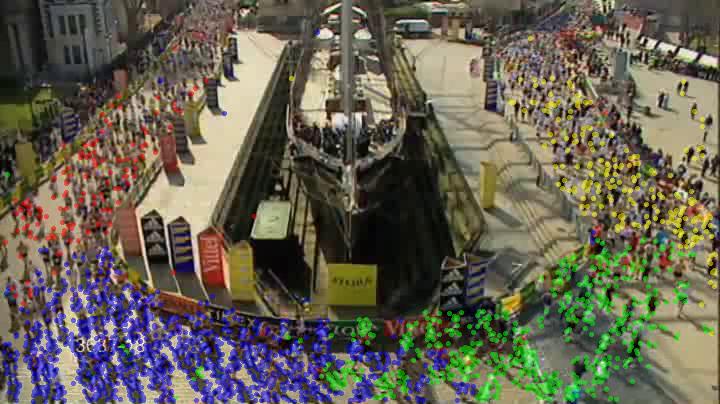}
        \caption{g} 
        \label{fig:3687SD61}
    \end{subfigure}
       \begin{subfigure}[t]{0.154\textwidth}
        \includegraphics[scale=0.1,width=0.8\textwidth]{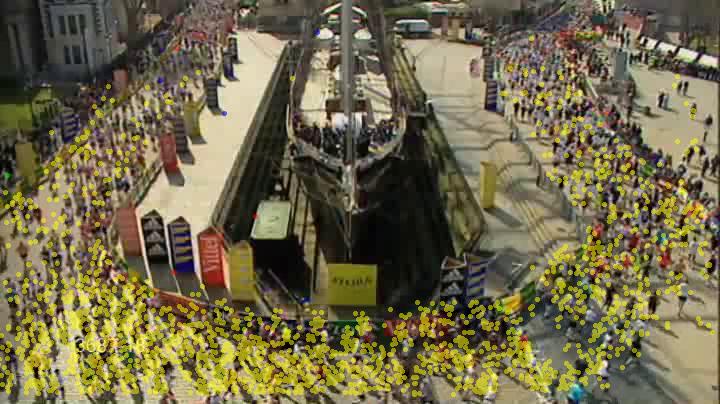}
        \caption{h} 
        \label{fig:3687SD62}
    \end{subfigure}
       \begin{subfigure}[t]{0.154\textwidth}
        \includegraphics[scale=0.1,width=0.8\textwidth]{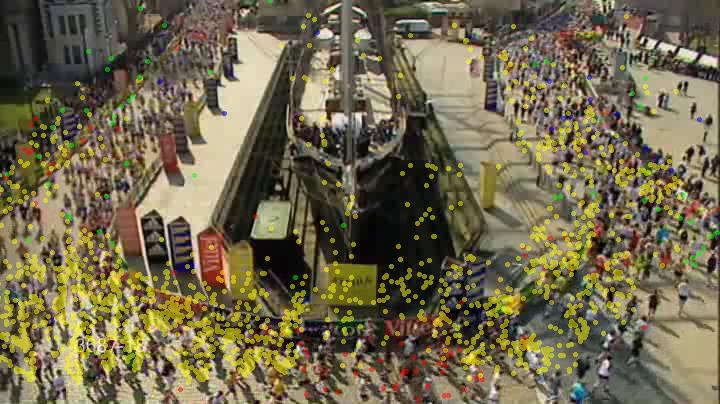}
        \caption{i} 
        \label{fig:3687SD63}
        \end{subfigure}
        \begin{subfigure}[t]{0.154\textwidth}
        \includegraphics[scale=0.1,width=0.8\textwidth]{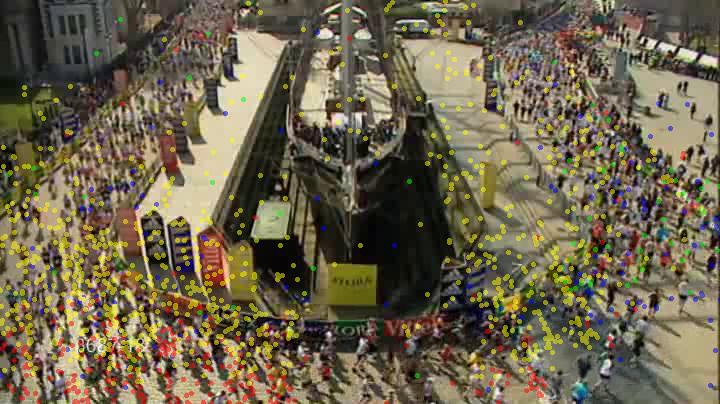}
        \caption{j} 
        \label{fig:3687AK61}
    \end{subfigure}
       \begin{subfigure}[t]{0.154\textwidth}
        \includegraphics[scale=0.1,width=0.8\textwidth]{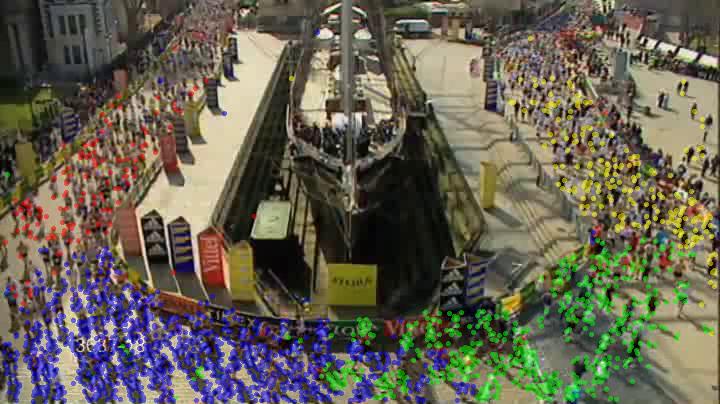}
        \caption{k} 
        \label{fig:3687AK62}
    \end{subfigure}
       \begin{subfigure}[t]{0.154\textwidth}
        \includegraphics[scale=0.1,width=0.8\textwidth]{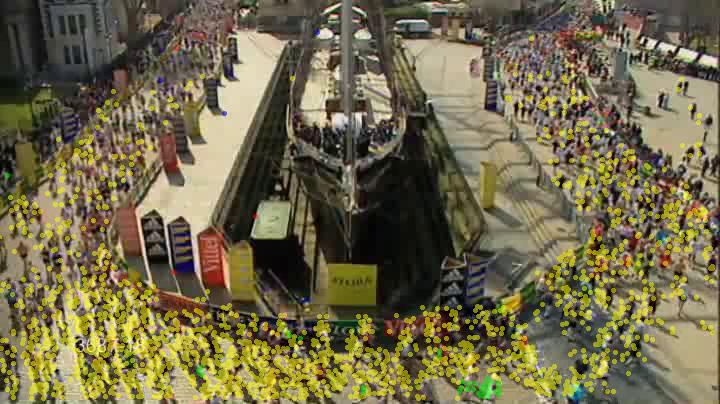}   
        \caption{l} 
        \label{fig:3687AK63}
        \end{subfigure}
            \begin{subfigure}[t]{0.154\textwidth}
        \includegraphics[scale=0.1,width=0.8\textwidth]{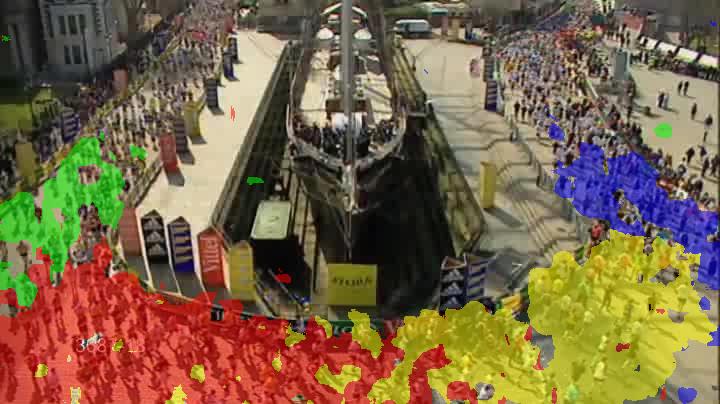}
        \caption{m} 
        \label{fig:DIHM368761}
        \end{subfigure}
               \begin{subfigure}[t]{0.154\textwidth}
        \includegraphics[scale=0.1,width=0.8\textwidth]{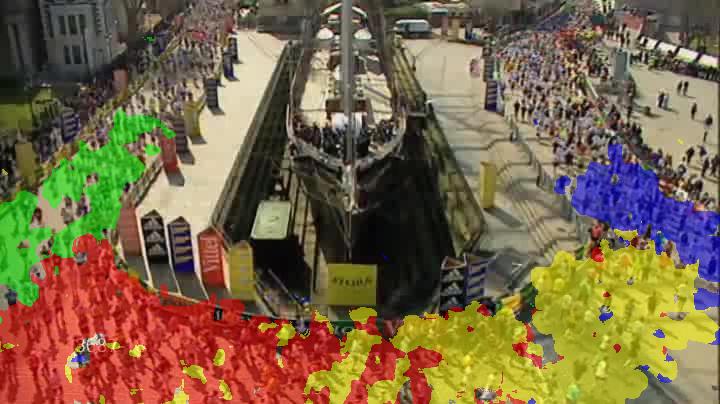}
        \caption{n} 
        \label{fig:DIHM368762}
        \end{subfigure}
               \begin{subfigure}[t]{0.154\textwidth}
        \includegraphics[scale=0.1,width=0.8\textwidth]{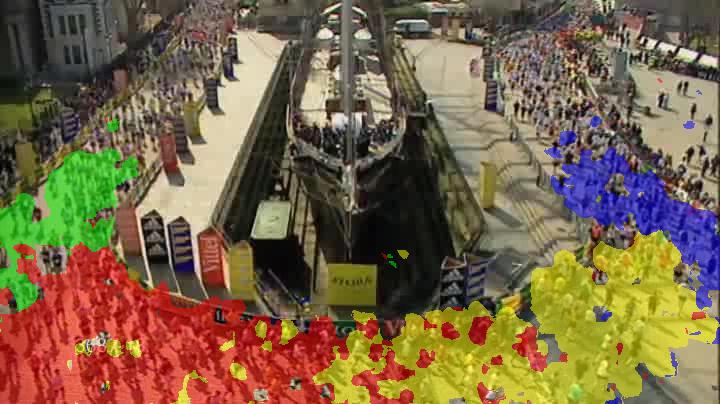}
        \caption{o} 
        \label{fig:DIHM368763}
        \end{subfigure}
\begin{subfigure}[t]{0.154\textwidth}
        \includegraphics[scale=1.0,width=0.8\textwidth]{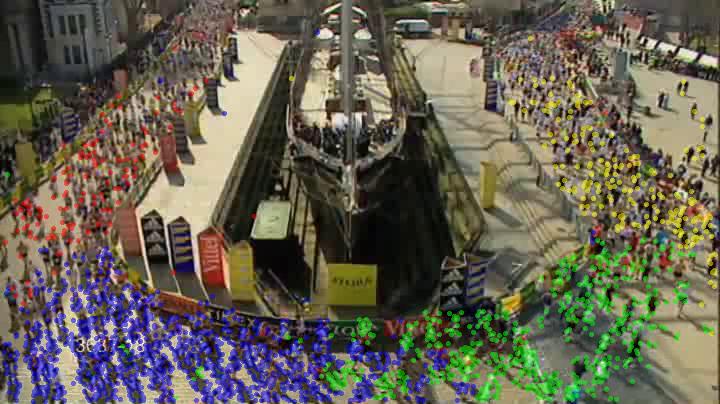}
        \caption{p} 
        \label{fig:ALE368761}
    \end{subfigure}
       \begin{subfigure}[t]{0.154\textwidth}
        \includegraphics[scale=0.1,width=0.8\textwidth]{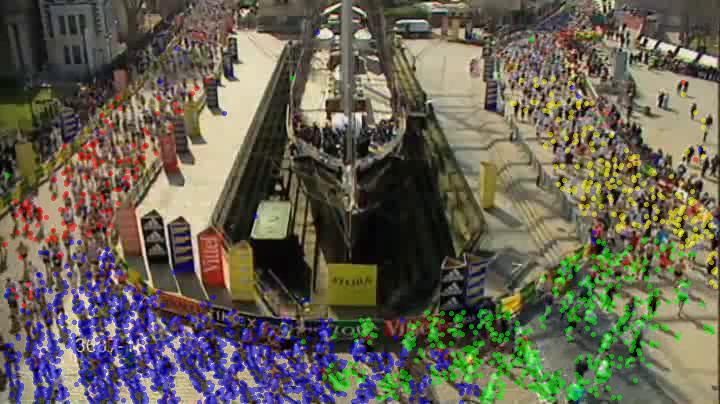}
        \caption{q} 
        \label{fig:ALE368762}
    \end{subfigure}
       \begin{subfigure}[t]{0.154\textwidth}
        \includegraphics[scale=0.1,width=0.8\textwidth]{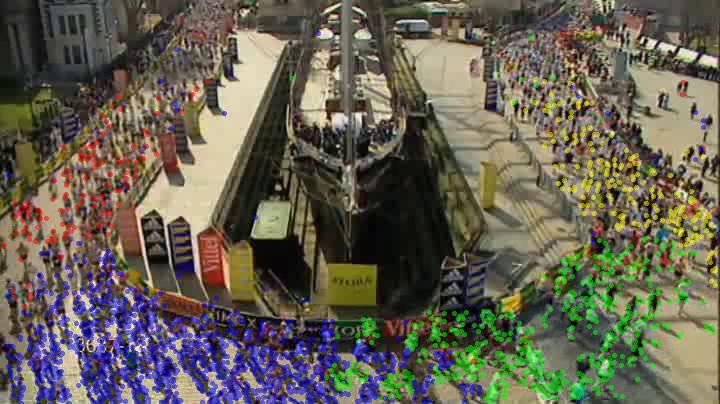}
        \caption{r} 
        \label{fig:ALE368763}
        \end{subfigure}
              \caption{($a$-$c$) Original Frames~(61-63) of the Marathon-III video, ($d$-$f$) Ground Truth Frames, ($g$-$i$) represent segmented outputs obtained using method proposed in~\cite{mahapatra2017transitions}, ($j$-$l$) represent segmentation outputs of  method~\cite{kulkarni2019sparse}, ($m$-$o$) represent segmentation outputs of method~\cite{ullah2017density}, and  ($p$-$r$) represent segmentation outputs of the proposed method, respectively. (Best viewed in color)}

\label{fig:3687}
\end{figure*}

\begin{figure*}[hbt!]     
   \centering
   \scriptsize
\begin{subfigure}[b]{0.4\textwidth}
\includegraphics[width=9.0 cm,height=6.0 cm]{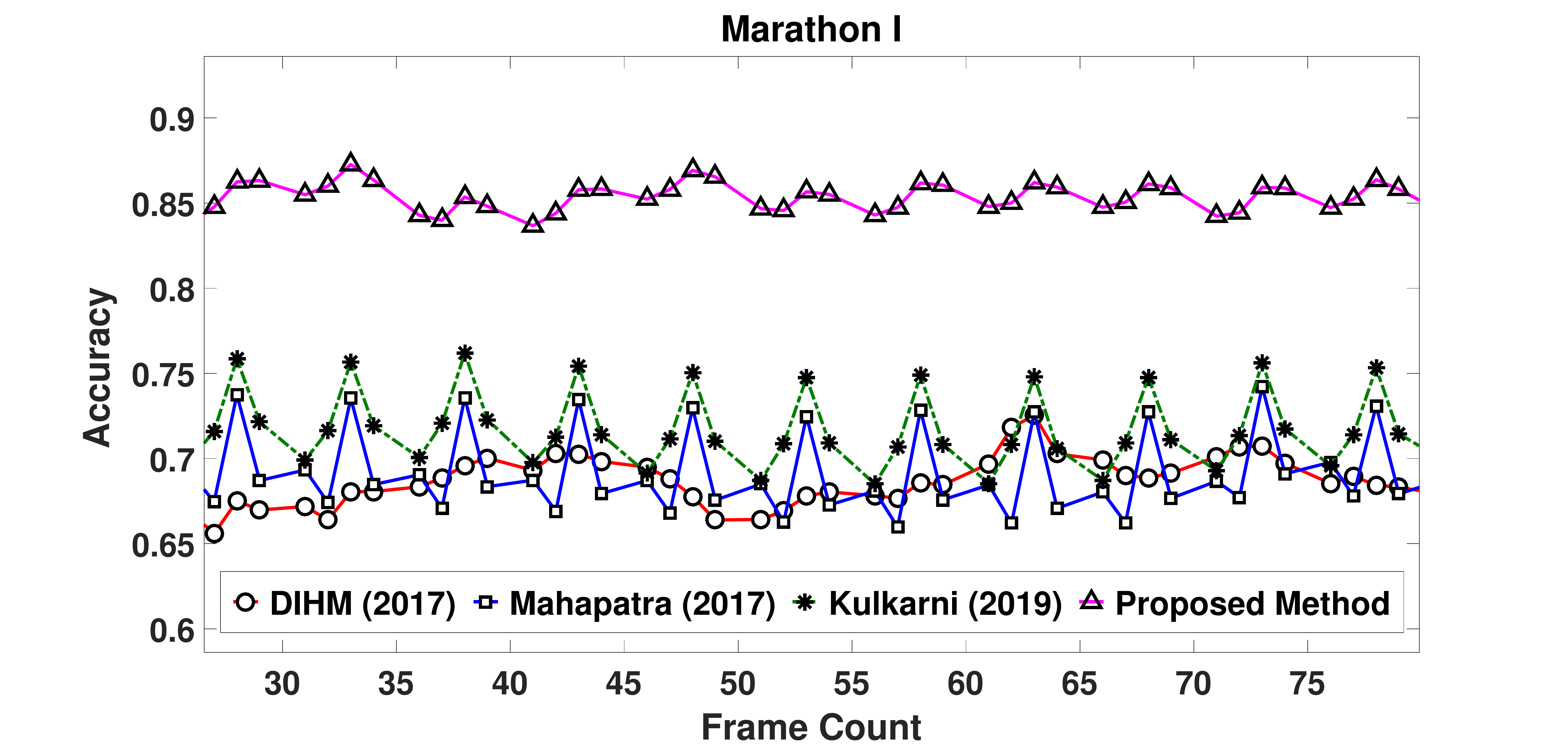}
\centering
        \caption{}
        \label{fig:accPlot688}
    \end{subfigure}
 \hspace{0.8cm}
       \begin{subfigure}[b]{0.4\textwidth}
\includegraphics[width=9.0 cm,height=6.0 cm]{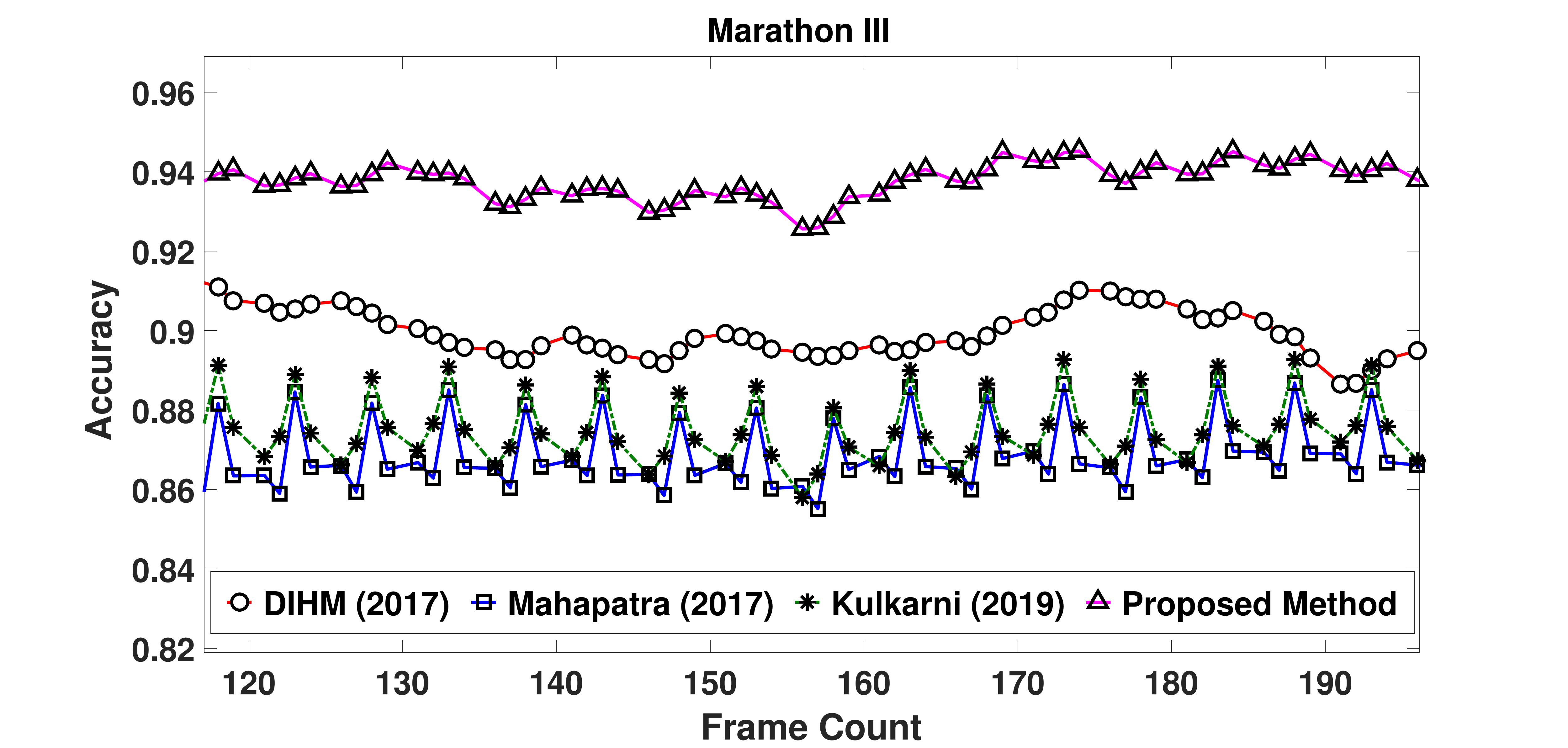} 
        \caption{}
        \label{fig:AccPlot3687}
    \end{subfigure}
    \hspace{0.8cm}    
       \begin{subfigure}[b]{0.4\textwidth}
        \includegraphics[width=9.0 cm,height=6.0 cm]{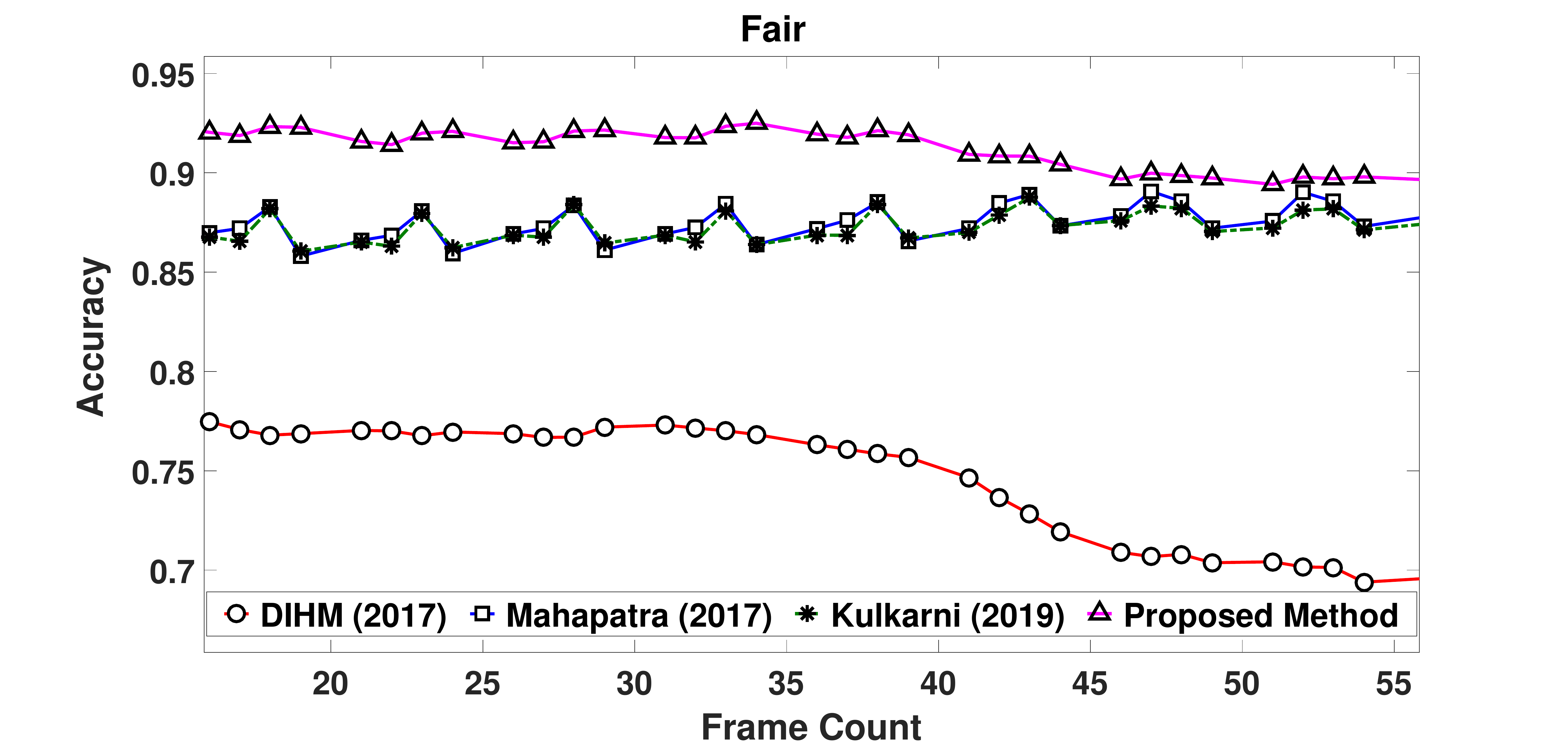}  
        \caption{}
        \label{fig:AccPlotFair}
        \end{subfigure}
         \hspace{1 cm}
       \begin{subfigure}[b]{0.4\textwidth}
        \includegraphics[width=9.0 cm,height=6.0 cm]{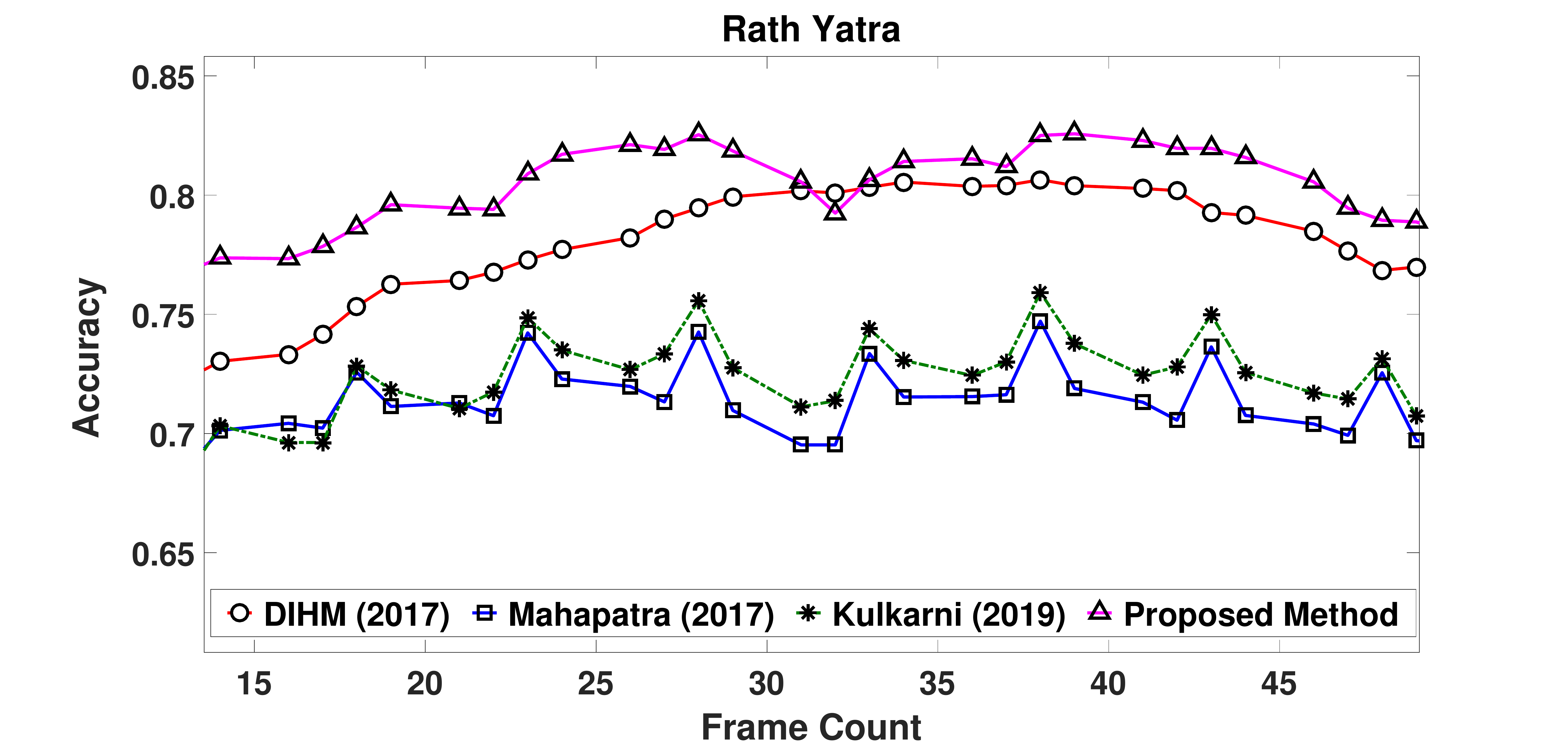} 
        \caption{}
        \label{fig:AccPlotRY}
        \end{subfigure}    
\caption{($a$-$d$) Frame-wise accuracy plot of various videos for the proposed method, \cite{mahapatra2017transitions}, \cite{kulkarni2019sparse} and \cite{ullah2017density}, respectively. (Best viewed in color)}
      \label{fig:aplots}
\end{figure*}

 In Marathon-I video, all people are running in one direction indicating it comprises of a unidirectional flow. The force models described in \cite{kulkarni2019sparse} and \cite{mahapatra2017transitions} have been implemented for comparisions. It has been observed that the exisitng models fail to accurately compute the flow vectors in terms of position and velocity. These can be observed in Figures~(\ref{fig:SD68816}-\ref{fig:SD68818}) and (\ref{fig:AK68816}-\ref{fig:AK68818}), respectively. On the contrary, the proposed segmentation scheme is able to compute the positions and velocities of the motion particles and thus segmenting the flow with better accuracy. The proposed method also outperforms the method proposed in \cite{ullah2017density}, where a hydrodynamics-based force model is used for segmentation. In the outputs generated by the hydrodynamics-based model in Figures (\ref{fig:DIHM368761}-\ref{fig:DIHM368763}), there are significant numbers of false-positives leading to poor accuracy as can be seen in the graphs shown in Figure\ref{fig:accPlot688}.

\par The Marathon-III video has elliptical motion. However, the flow comprises of four directions indicated by different colors as seen in the ground-truth images in Figures (\ref{fig:GT368761}-\ref{fig:GT368763}). The proposed method is able to segment these multi-directional flows with an accuracy $93.11\%$ that is better than the force models described in \cite{mahapatra2017transitions} and \cite{kulkarni2019sparse}. The hydrodynamics model~\cite{ullah2017density} segments the multi-directional flows. However, there are some over-segmentations that can be observed in Figures~(\ref{fig:DIHM368761}-\ref{fig:DIHM368763}).
 \par The Fair video is a crowd mixing video with two dominant flows moving in opposite directions. The proposed method is able to segment these flows. However, the force models proposed in \cite{kulkarni2019sparse} and \cite{mahapatra2017transitions} segment them as a unidirectional flow. The hydrodynamics-based model segments these flows with more false-positives.
\begin{figure*}[hbt!]     
    \centering
    \captionsetup[subfigure]{labelformat=empty}
\begin{subfigure}[t]{0.155\textwidth}
        \includegraphics[scale=0.1,width=0.8\textwidth]{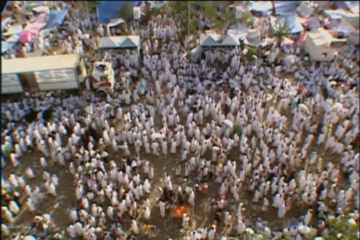}
        \caption{a} 
        \label{fig:OriginalFair41}
    \end{subfigure}
       \begin{subfigure}[t]{0.155\textwidth}
        \includegraphics[scale=0.1,width=0.8\textwidth]{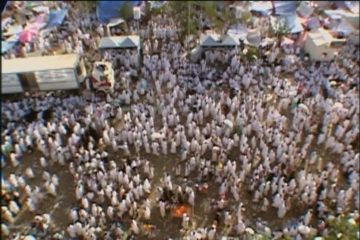}
        \caption{b} 
        \label{fig:OriginalFair42}
    \end{subfigure}
       \begin{subfigure}[t]{0.155\textwidth}
        \includegraphics[scale=0.1,width=0.8\textwidth]{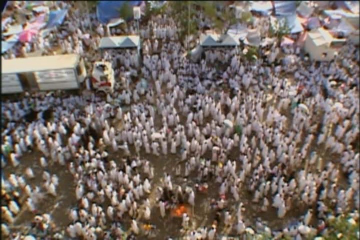}
        \caption{c} 
        \label{fig:OriginalFair43}
        \end{subfigure}
          \begin{subfigure}[t]{0.155\textwidth}
        \includegraphics[scale=0.1,width=0.8\textwidth]{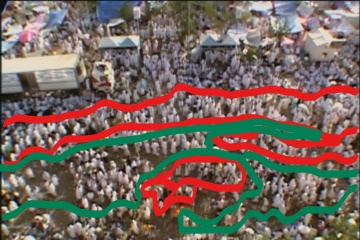}
        \caption{d} 
        \label{fig:GTFair41}
    \end{subfigure}
       \begin{subfigure}[t]{0.155\textwidth}
        \includegraphics[scale=0.1,width=0.8\textwidth]{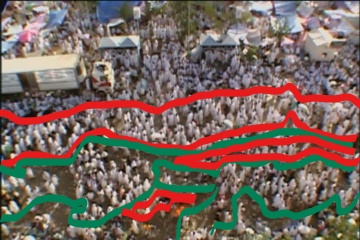}
        \caption{e} 
        \label{fig:GTFair42}
    \end{subfigure}
       \begin{subfigure}[t]{0.155\textwidth}
        \includegraphics[scale=0.1,width=0.8\textwidth]{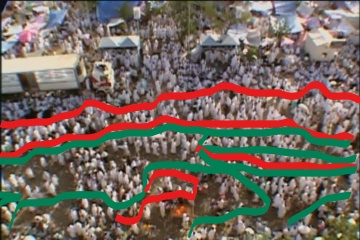}
        \caption{f} 
        \label{fig:GTFair43}
        \end{subfigure}
                   \begin{subfigure}[t]{0.155\textwidth}
        \includegraphics[scale=0.1,width=0.8\textwidth]{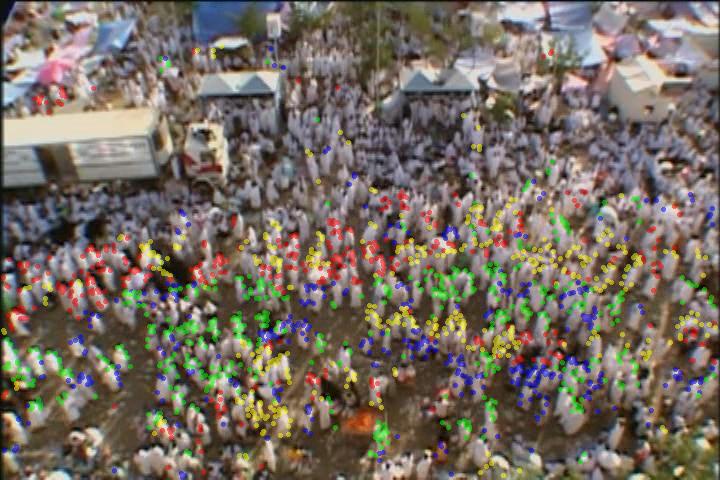}
        \caption{g} 
        \label{fig:FairSD41}
    \end{subfigure}
       \begin{subfigure}[t]{0.155\textwidth}
        \includegraphics[scale=0.1,width=0.8\textwidth]{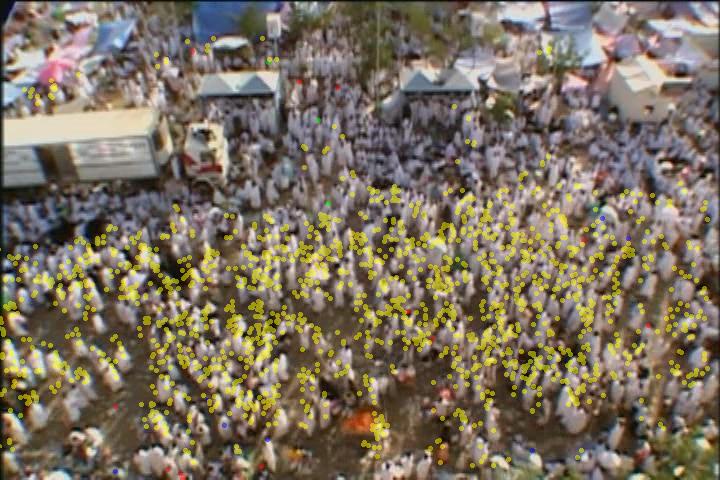}
        \caption{h} 
        \label{fig:FairSD42}
    \end{subfigure}
       \begin{subfigure}[t]{0.155\textwidth}
        \includegraphics[scale=0.1,width=0.8\textwidth]{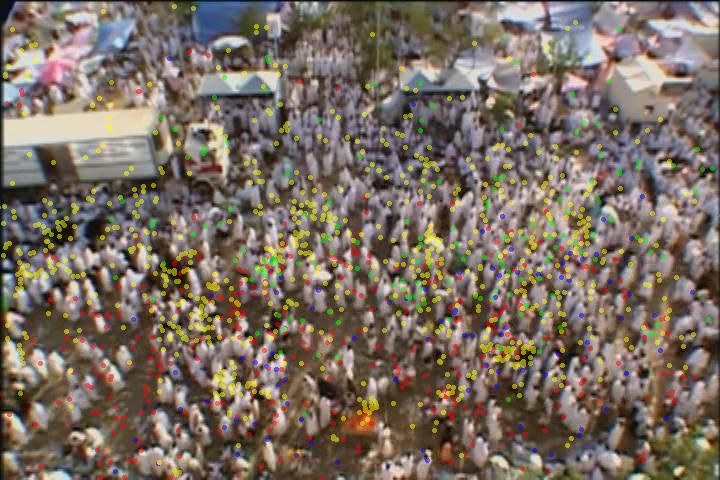}
        \caption{i} 
        \label{fig:FairSD43}
        \end{subfigure}
        \begin{subfigure}[t]{0.155\textwidth}
        \includegraphics[scale=0.1,width=0.8\textwidth]{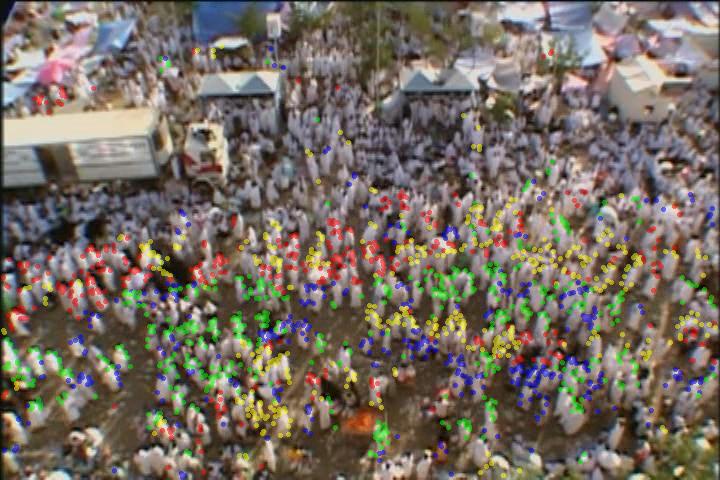}
        \caption{j} 
        \label{fig:FairAK41}
    \end{subfigure}
       \begin{subfigure}[t]{0.155\textwidth}
        \includegraphics[scale=0.1,width=0.8\textwidth]{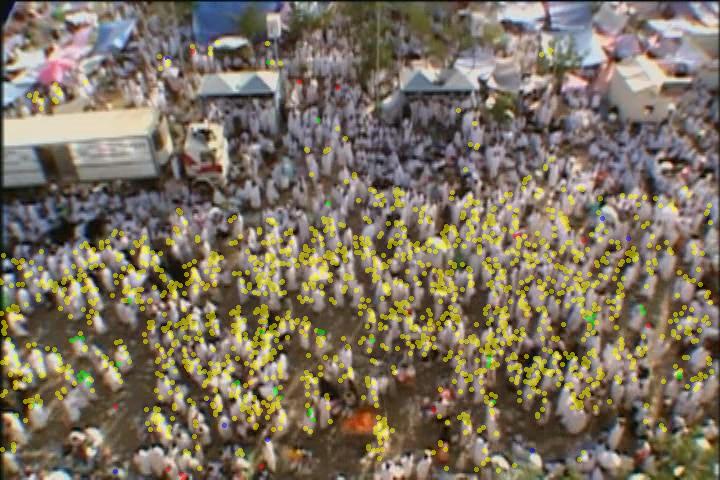}
        \caption{k} 
        \label{fig:FairAK42}
    \end{subfigure}
       \begin{subfigure}[t]{0.155\textwidth}
        \includegraphics[scale=0.1,width=0.8\textwidth]{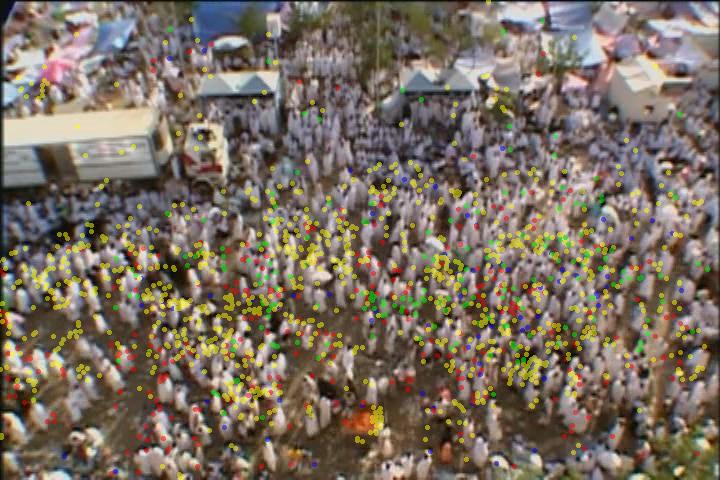}   
        \caption{l} 
        \label{fig:FairAK43}
        \end{subfigure}
            \begin{subfigure}[t]{0.155\textwidth}
        \includegraphics[scale=0.1,width=0.8\textwidth]{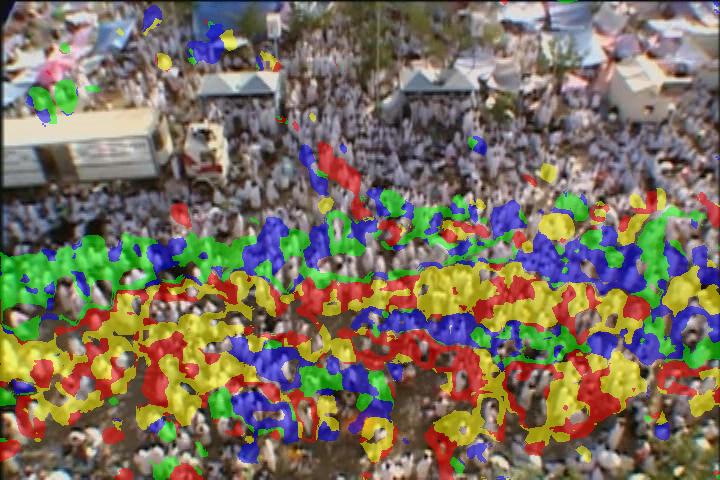}
        \caption{m} 
        \label{fig:DIHMFair41}
        \end{subfigure}
               \begin{subfigure}[t]{0.155\textwidth}
        \includegraphics[scale=0.1,width=0.8\textwidth]{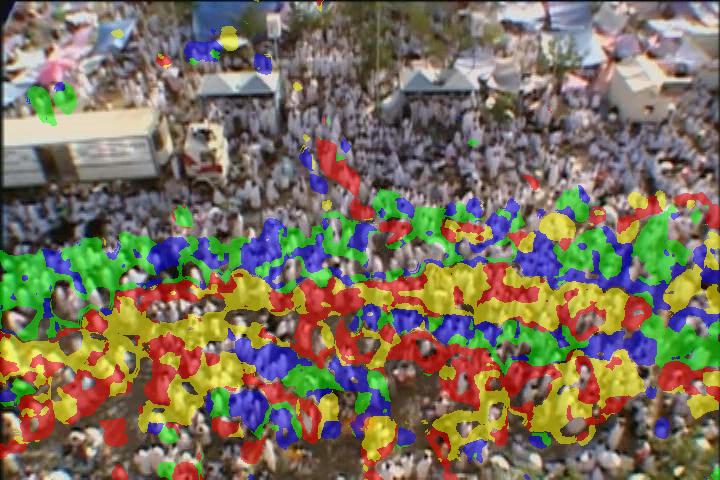}
        \caption{n} 
        \label{fig:DIHMFair42}
        \end{subfigure}
               \begin{subfigure}[t]{0.155\textwidth}
        \includegraphics[scale=0.1,width=0.8\textwidth]{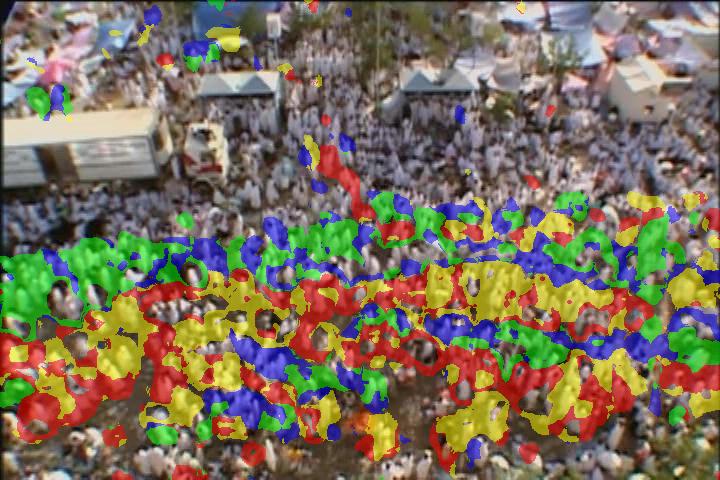}
        \caption{o} 
        \label{fig:DIHMFair43}
        \end{subfigure}
\begin{subfigure}[t]{0.155\textwidth}
        \includegraphics[scale=1.0,width=0.8\textwidth]{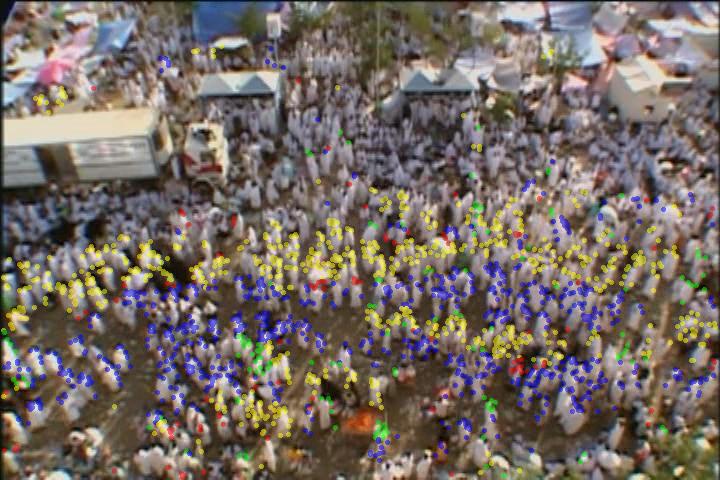}
        \caption{p} 
        \label{fig:ALEFair41}
    \end{subfigure}
       \begin{subfigure}[t]{0.155\textwidth}
        \includegraphics[scale=0.1,width=0.8\textwidth]{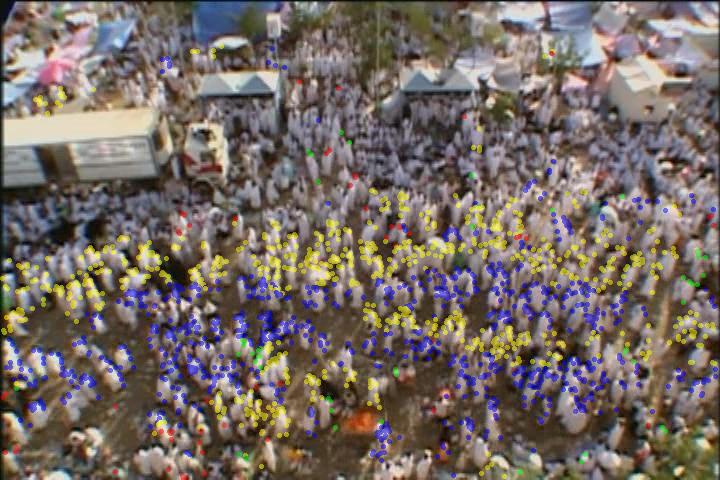}
        \caption{q} 
        \label{fig:ALEFair42}
    \end{subfigure}
       \begin{subfigure}[t]{0.155\textwidth}
        \includegraphics[scale=0.1,width=0.8\textwidth]{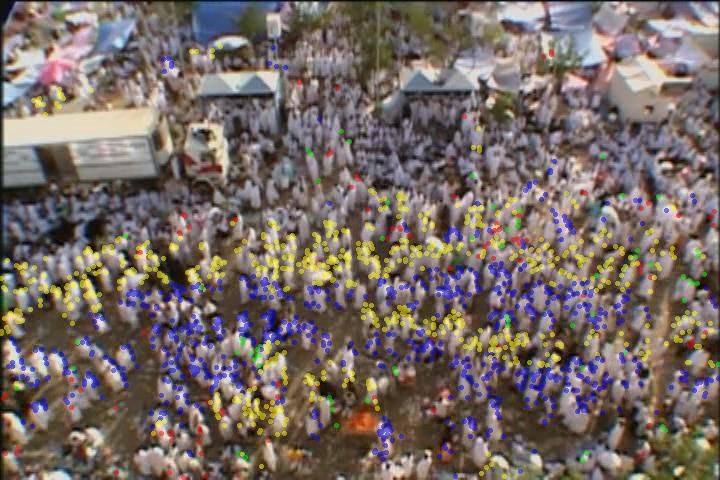}
       \caption{r} 
        \label{fig:ALEFair43}
        \end{subfigure}
          \caption{($a$-$c$) Original Frames~(41-43) of the Fair video, ($d$-$f$) Ground Truth Frames, ($g$-$i$) represent segmented outputs obtained using method proposed in~\cite{mahapatra2017transitions}, ($j$-$l$) represent segmentation outputs of  method~\cite{kulkarni2019sparse}, ($m$-$o$) represent segmentation outputs of method~\cite{ullah2017density}, and  ($p$-$r$) represent segmentation outputs of the proposed method, respectively. (Best viewed in color)}
\label{fig:Fair}
\end{figure*}
\par In the Rath Yatra video, both crowd mixing and cart pulling event can be observed. The cart pulling is the dominant flow movement in the video. The proposed method is able to segment this dominant flow (in red color) and it is able to segment other flows (in blue color). The force models in \cite{kulkarni2019sparse} and~\cite{mahapatra2017transitions} fail to segment these flows. Moreover, the estimated directions are not consistent. The hydrodynamics-based model \cite{ullah2017density} fails to segment the dominant flows properly. The average frame accuracy for this video using the proposed method has been found to be $90.56\%$.
\begin{figure*}[hbt!]    
   \centering
   \captionsetup[subfigure]{labelformat=empty}
\begin{subfigure}[t]{0.15\textwidth}
        \includegraphics[scale=0.1,width=0.8\textwidth]{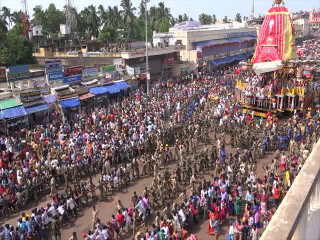}
        \caption{a} 
        \label{fig:OriginalRY31}
    \end{subfigure}
       \begin{subfigure}[t]{0.15\textwidth}
        \includegraphics[scale=0.1,width=0.8\textwidth]{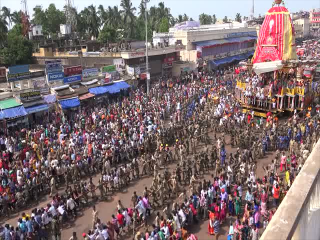}
        \caption{b} 
        \label{fig:OriginalRY32}
    \end{subfigure}
       \begin{subfigure}[t]{0.15\textwidth}
        \includegraphics[scale=0.1,width=0.8\textwidth]{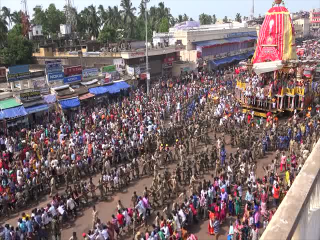}
        \caption{c} 
        \label{fig:OriginalRY33}
        \end{subfigure}
          \begin{subfigure}[t]{0.15\textwidth}
        \includegraphics[scale=0.1,width=0.8\textwidth]{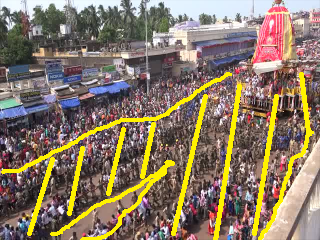}
        \caption{d} 
        \label{fig:GTRY31}
    \end{subfigure}
       \begin{subfigure}[t]{0.15\textwidth}
        \includegraphics[scale=0.1,width=0.8\textwidth]{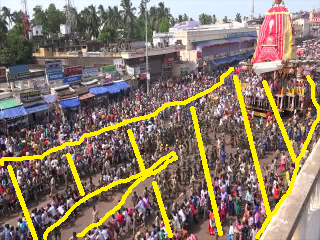}
        \caption{e} 
        \label{fig:GTRY32}
    \end{subfigure}
       \begin{subfigure}[t]{0.15\textwidth}
        \includegraphics[scale=0.1,width=0.8\textwidth]{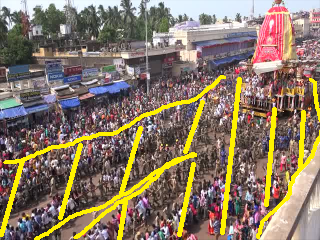}
        \caption{f} 
        \label{fig:GTRY33}
        \end{subfigure}
                   \begin{subfigure}[t]{0.15\textwidth}
        \includegraphics[scale=0.1,width=0.8\textwidth]{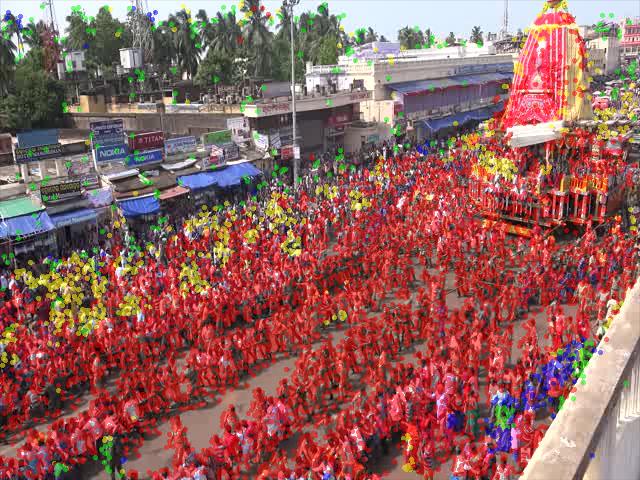}
        \caption{g} 
        \label{fig:RY1SD31}
    \end{subfigure}
       \begin{subfigure}[t]{0.15\textwidth}
        \includegraphics[scale=0.1,width=0.8\textwidth]{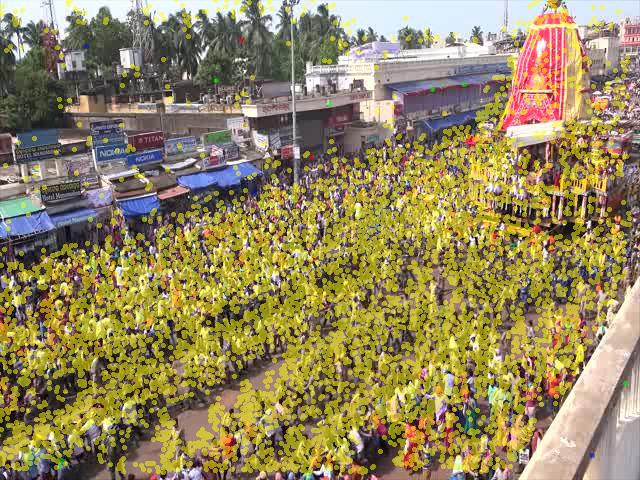}
        \caption{h} 
        \label{fig:RY1SD32}
    \end{subfigure}
       \begin{subfigure}[t]{0.15\textwidth}
        \includegraphics[scale=0.1,width=0.8\textwidth]{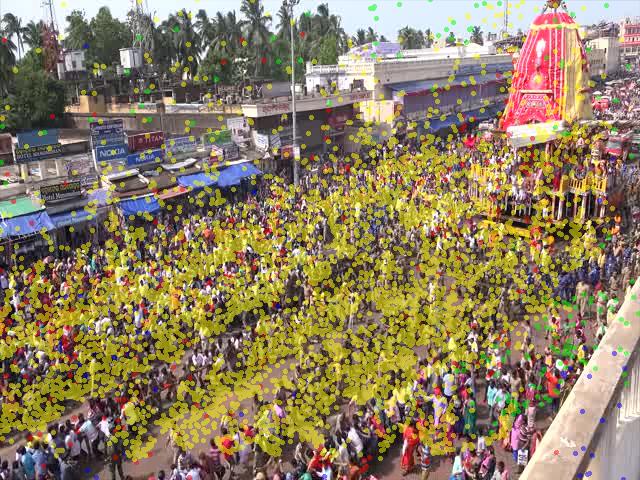}
        \caption{i} 
        \label{fig:RY1SD33}
        \end{subfigure}
        \begin{subfigure}[t]{0.15\textwidth}
        \includegraphics[scale=0.1,width=0.8\textwidth]{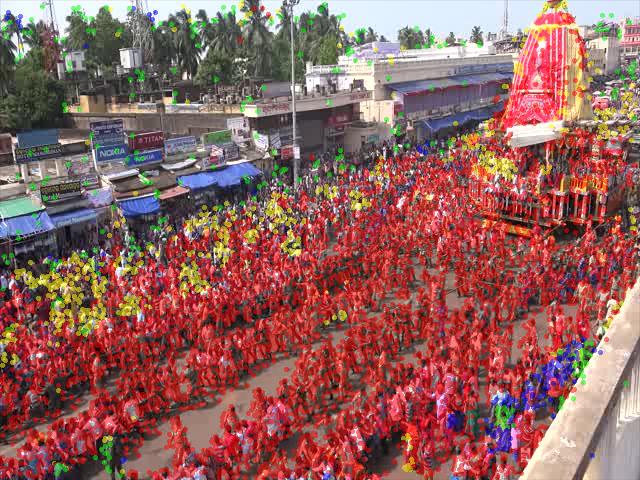}
        \caption{j} 
        \label{fig:RY1AK31}
    \end{subfigure}
       \begin{subfigure}[t]{0.15\textwidth}
        \includegraphics[scale=0.1,width=0.8\textwidth]{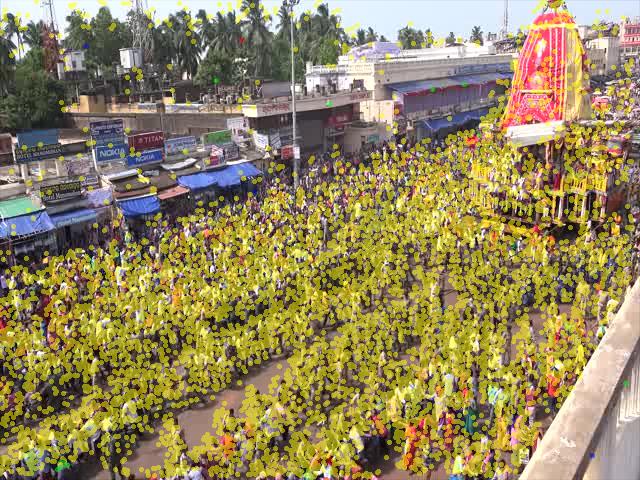}
        \caption{k} 
        \label{fig:RY1AK32}
    \end{subfigure}
       \begin{subfigure}[t]{0.15\textwidth}
        \includegraphics[scale=0.1,width=0.8\textwidth]{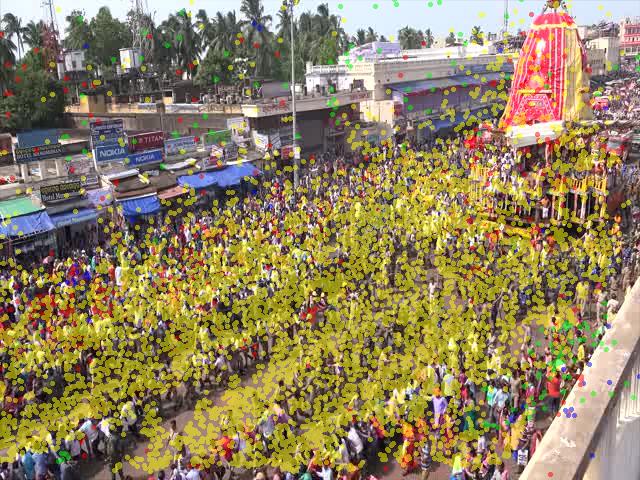}   
        \caption{l} 
        \label{fig:RY1AK33}
        \end{subfigure}
            \begin{subfigure}[t]{0.15\textwidth}
        \includegraphics[scale=0.1,width=0.8\textwidth]{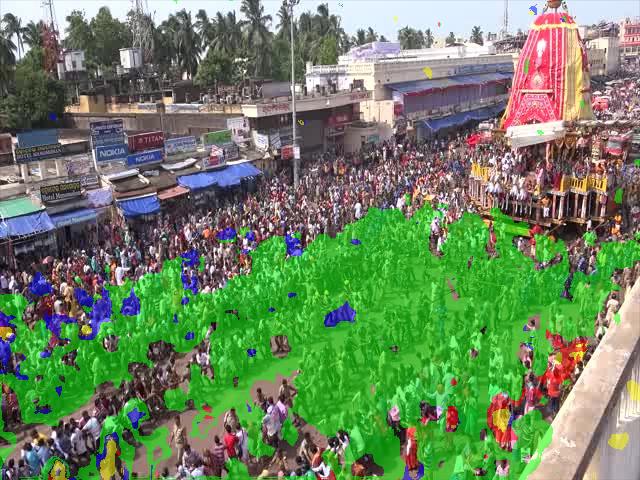}
        \caption{m} 
        \label{fig:DIHMRY131}
        \end{subfigure}
               \begin{subfigure}[t]{0.15\textwidth}
        \includegraphics[scale=0.1,width=0.8\textwidth]{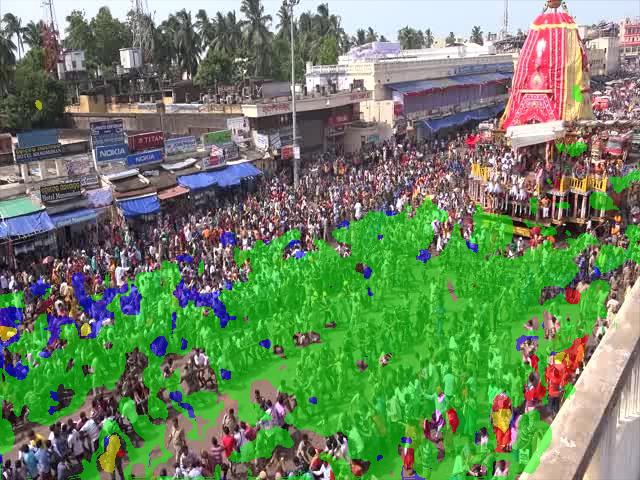}
       \caption{n} 
        \label{fig:DIHMRY132}
        \end{subfigure}
               \begin{subfigure}[t]{0.15\textwidth}
        \includegraphics[scale=0.1,width=0.8\textwidth]{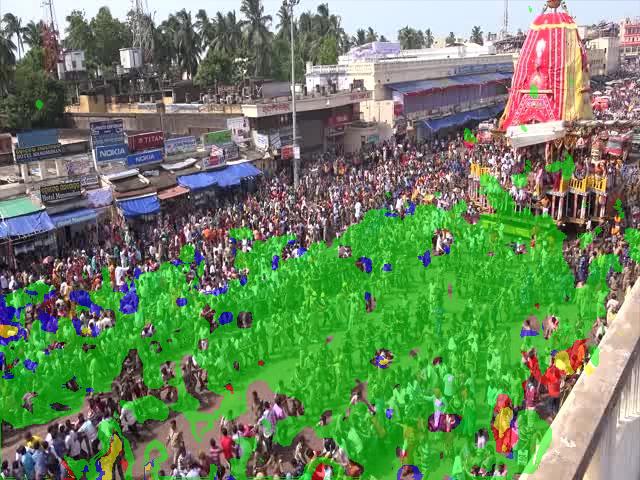}
        \caption{o} 
        \label{fig:DIHMRY133}
        \end{subfigure}
\begin{subfigure}[t]{0.15\textwidth}
        \includegraphics[scale=1.0,width=0.8\textwidth]{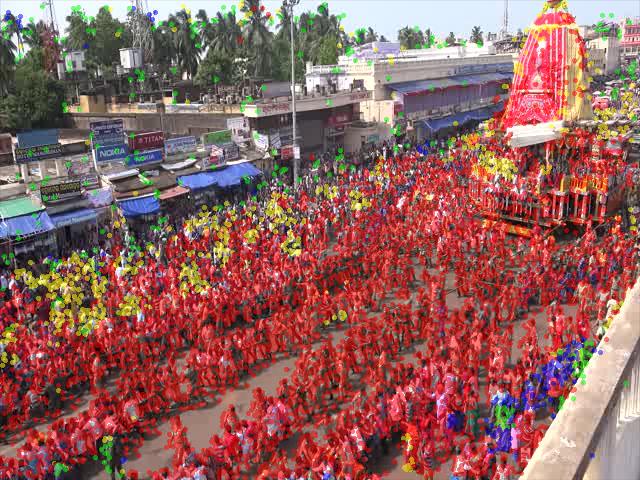}
        \caption{p} 
        \label{fig:ALERY131}
    \end{subfigure}
       \begin{subfigure}[t]{0.15\textwidth}
        \includegraphics[scale=0.1,width=0.8\textwidth]{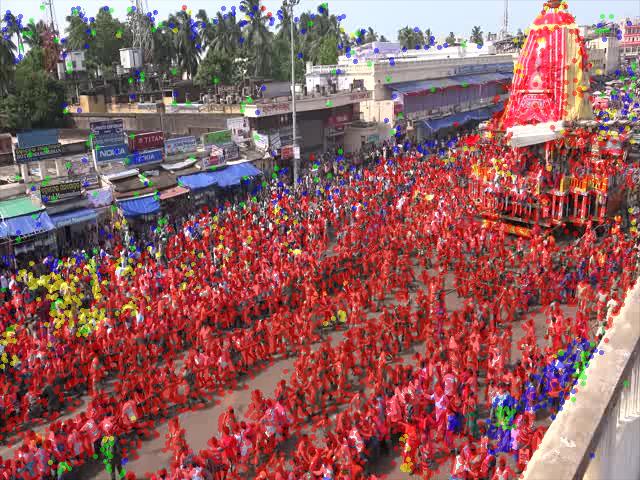}
        \caption{q} 
        \label{fig:ALERY132}
    \end{subfigure}
       \begin{subfigure}[t]{0.15\textwidth}
        \includegraphics[scale=0.1,width=0.8\textwidth]{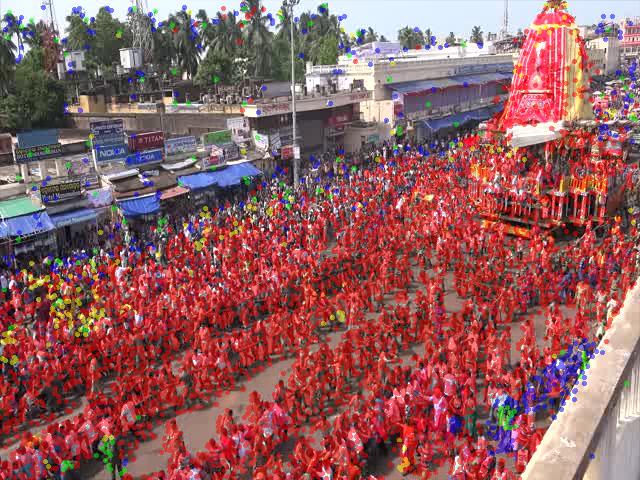}
        \caption{r} 
        \label{fig:ALERY133}
        \end{subfigure}
     \caption{($a$-$c$) Original recorded Frames~(31-33) of the Rath Yatra-I video,  ($d$-$f$) Ground Truth Frames, ($g$-$i$) represent segmented outputs obtained using method proposed in~\cite{mahapatra2017transitions}, ($j$-$l$) represent segmentation outputs of  method~\cite{kulkarni2019sparse}, ($m$-$o$) represent segmentation outputs of method~\cite{ullah2017density}, and  ($p$-$r$) represent segmentation outputs of the proposed method, respectively. (Best viewed in color)}
\label{fig:RY}
\end{figure*}
\par The average accuracies for all methods are summarized in Table~\ref{table:accuracy}. The accuracy/frame plots of all methods for various videos are represented in Figure \ref{fig:aplots}.
\begin{table}[]
\centering
\caption{Comparison of the proposed method with state-of-the-art in terms of accuracy}
\label{table:accuracy}
\scriptsize
\resizebox{0.47\textwidth}{!}{%
\begin{tabular}{|l|c|c|c|c|}
\hline
\multicolumn{1}{|c|}{\multirow{2}{*}{\#Dataset}} & \multicolumn{4}{c|}{Accuracy (in \%)} \\ \cline{2-5} 
\multicolumn{1}{|c|}{} & Proposed Method & \multicolumn{1}{l|}{\cite{mahapatra2017transitions}} & \multicolumn{1}{l|}{\cite{kulkarni2019sparse}} & \multicolumn{1}{l|}{\cite{ullah2017density}} \\ \hline
Marathon-I & 82.89 & 66.59 & 69.03 & 67.71 \\ \hline
Marathon-III & 93.11 & 85.54 & 86.61 & 90.37 \\ \hline
Fair & 90.56 & 86.03 & 86.23 & 74.46 \\ \hline
Rath Yatra & 78.48 & 68.58 & 69.42 & 77.12 \\ \hline
\end{tabular}%
}
\end{table}
\par
\begin{figure}[hbt!]
	\centering	
    \includegraphics[width=9.0 cm,height=5.0 cm]{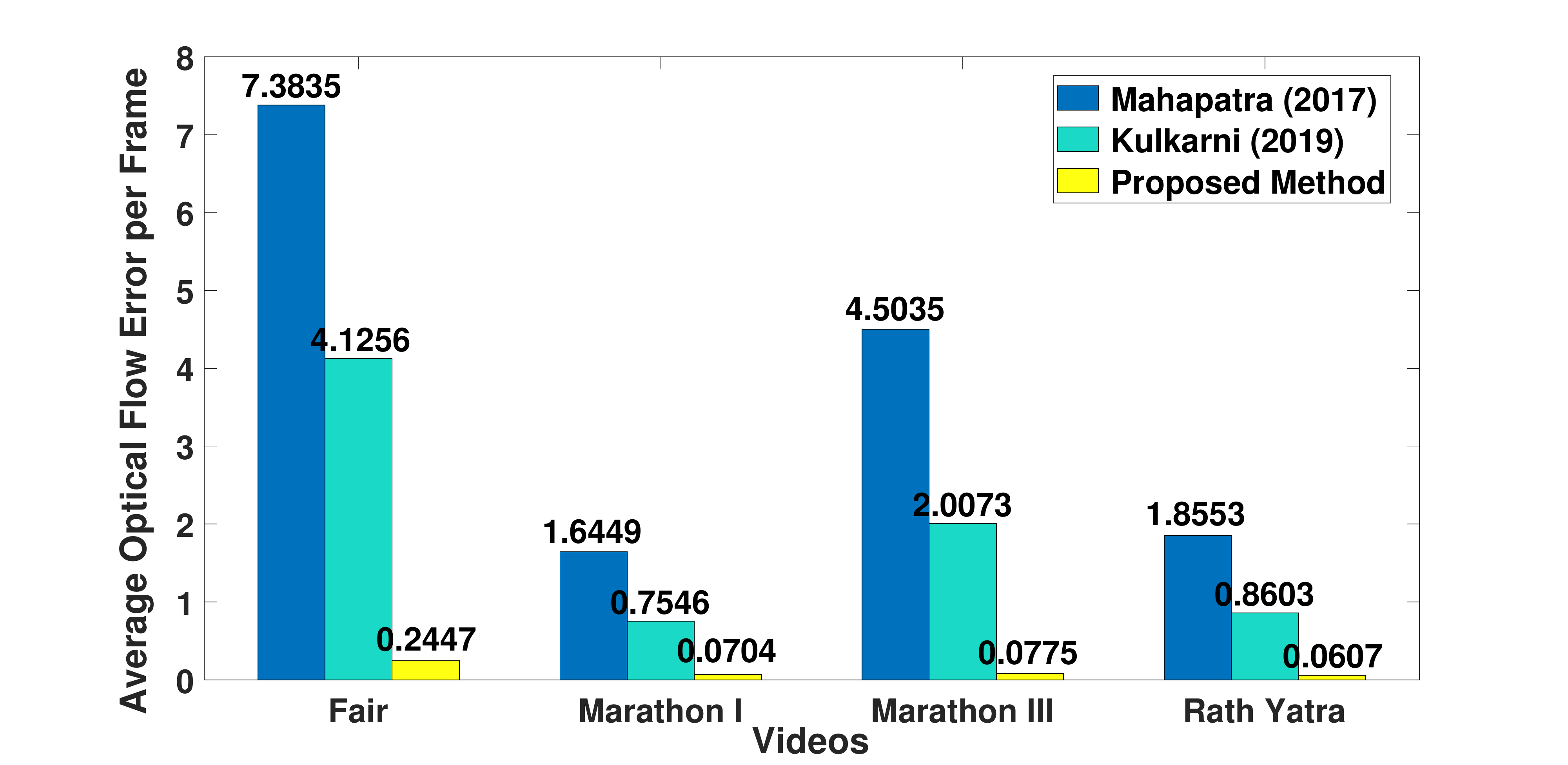}
    \caption{Frame-wise average optical error per frame bar graph plot of various videos for the proposed method, \cite{mahapatra2017transitions}, and \cite{kulkarni2019sparse}, respectively. (Best viewed in color)}  
    \label{fig:errorOF}
\end{figure}

 The proposed force model and the force models in \cite{kulkarni2019sparse} and \cite{mahapatra2017transitions} have also been compared with the optical flow baselines. The computed positions and velocities of the particles are compared with the  positions and velocities computed using optical flow computed for $t+1$ and $t+2$ frames by computing the average optical flow error between them and plotting them for every frame in the video. The errors for all the models for all videos are plotted in Figure \ref{fig:errorOF}. It may be  observed that the average optical flow error/frame for the proposed method is lesser as compared to other physics-based models.
\section{Conclusion and Future Scope}
\label{sec:Conclusions}
In this work, an approach based on active Langevin equation has been used to understand the motion flows in crowd videos. The active Langevin equation models the motion particles in crowd similar to the colloidal particles moving in the fluid. The segmentation scheme based on this model segments straight line motion as well as curvy motion with notable accuracy. The usage of windowing scheme ensures a significant decrease in the number of computations as optical flow is calculated for two consecutive frames of the window and for the remaining frames, the proposed force model computes the flow positions and velocities to obtain temporal segmentation. In the future, the proposed model can be augmented with machine-learning approaches for identifying and predicting abnormal regions in crowded scenes. 
\section*{Acknowledgment}
The authors are grateful to Science and Engineering Research Board (SERB), Department of Science and Technology, Government of India for funding this research work through the grant YSS/2014/000046.

\label{sec:ack}
\bibliographystyle{IEEEtran}      

\end{document}